\documentclass[letterpaper]{article}
\usepackage[ margin= 0.993 in]{geometry}
\usepackage{graphicx} % DO NOT CHANGE THIS
\usepackage{caption} % DO NOT CHANGE THIS AND DO NOT ADD ANY OPTIONS TO IT
\usepackage{mathtools}
\usepackage{verbatim}
\usepackage{amsmath}
\usepackage{amsfonts}
\usepackage{float}
\usepackage{subfigure}
\usepackage{algorithm}
\usepackage[noend]{algpseudocode}
\usepackage{booktabs}
\usepackage{enumitem}
\usepackage{xparse, etoolbox, mathabx}
\usepackage[colorlinks=true,urlcolor=blue]{hyperref}
\usepackage{hyperref}

\usepackage{authblk}
\usepackage{microtype}
\usepackage{graphicx}
\usepackage{subfigure}
\usepackage{booktabs} % for professional tables
\usepackage{float}
\usepackage{graphicx}
\usepackage{amsfonts}
\usepackage{bbm}
\usepackage[colorlinks=true,urlcolor=blue]{hyperref}
\usepackage{color}
\usepackage{booktabs}
\usepackage{url}

\usepackage{amsmath}
\usepackage{amssymb}
\usepackage{mathtools}
\usepackage{amsthm}
\usepackage{hyperref}

\usepackage{hyperref}

\newcommand{\rmifnextchar}[3]{%
  \begingroup
  \ltx@LocToksA{\endgroup#2}%
  \ltx@LocToksB{\endgroup#3}%
  \ltx@ifnextchar{#1}{%
    \def\next{\the\ltx@LocToksA}%
    \afterassignment\next
    \let\scratch= %
  }{%
    \the\ltx@LocToksB
  }%
}

\usepackage{hyperref}
\newcommand{\vct}[1]{\boldsymbol{#1}}
\newcommand{\vx}{\vct{x}}
\newcommand{\vy}{\vct{y}}
\newcommand{\va}{\vct{a}}
\newcommand{\vb}{\vct{b}}

\newcommand{\vdelta}{\boldsymbol{\delta}}
\newcommand{\valpha}{\boldsymbol{\alpha}}
\newcommand{\vbeta}{\boldsymbol{\beta}}

\newcommand{\R}{\mathbb{R}}

\newcommand{\mtx}[1]{\boldsymbol{#1}}

\newcommand{\mH}{\mtx{H}}
\newcommand{\mX}{\mtx{X}}
\newcommand{\mY}{\mtx{Y}}
\newcommand{\mM}{\mtx{M}}
\newcommand{\mP}{\mtx{P}}
\newcommand{\mC}{\mtx{C}}
\newcommand{\mL}{\mtx{L}}
\newcommand{\mI}{\mtx{I}}

\newtheorem{proposition}{Proposition}[section]
\newtheorem{corollary}{Corollary}[section]

\title{Learning Sinkhorn divergences for supervised change point detection}

\author{Nauman Ahad, Eva L. Dyer, Keith B. Hengen, Yao Xie, Mark A. Davenport}
\begin{document}
\maketitle

\begin{abstract}
Many modern applications require detecting change points in complex sequential data.
Most existing methods for change point detection are unsupervised and, as a consequence, lack any information regarding what kind of changes we want to detect or if some kinds of changes are safe to ignore. This often results in poor change detection performance.
%Though examples of true change point instances are often available, there is only limited work that explores how these instances can be used to improve change point detection. 
We present a novel change point detection framework that uses true change point instances as supervision for learning a ground metric such that Sinkhorn divergences can be then used
%Available change points labels cab be used to obtain triplets sub-sequences that can be used to learn this ground metric. 
%This learned metric can be used by Sinkhorn divergences %
in two-sample tests on sliding windows to detect change points in an online manner. Our method can be used to learn a sparse metric which can be useful for both feature selection and interpretation in high-dimensional change point detection settings.
Experiments on simulated as well as real world sequences show that our proposed method can substantially improve change point detection performance over existing unsupervised change point detection methods using only few labeled change point instances.
\end{abstract}
{\let\thefootnote\relax\footnote{{N.Ahad, E.L.Dyer and M.A.Davenport are with the School of Electrical and Computer Engineering, Georgia Tech, Atlanta, GA,  USA.
K.B.Hengen is with the Department of Biology, Washington University in St. Louis, St. Louis, USA.
Y.Xie is with the School of Industrial and Systems Engineering, Georgia Tech, Atlanta, GA, USA. This work was supported, in part, by NSF awards DMS-2134037, IIS-2039741, NIH award 1R01EB029852-01, and a generous gift from the McKnight Foundation.\\
Correspondence to nahad3@gatech.edu, mdav@gatech.edu}} 

\section{Introduction}

Sequential data permeates our daily lives. 
 %Kitchen appliances,Sleep and fitness trackers are just some of the numerous devices that analyze sequential data. 
 Many applications, ranging from health care to climate monitoring, require detecting change points in sequential data
 \cite{beaulieu2012change,liu2018change}.
Such applications often involve increasingly complex and high-dimensional data, leading to challenging change point detection problems.

More precisely, let $\mX$ denote a sequence $\vx_1, \vx_2, \ldots,\vx_t \in \mathbb{R}^d$. We say that $\mX$ has a change point at index $n_c$ if $\vx_{n_c}, \vx_{n_c+1}, \ldots$ are generated according to a different distribution from $\vx_{n_c-1}, \vx_{n_c-2}, \ldots$. The problem of change point detection is to identify the indices corresponding to such change points.  There is a rich literature studying this problem, with approaches that can be broadly classified as either \emph{offline} methods that focus on partitioning a complete sequence \cite{truong2020selective} or \emph{online} methods that that can operate in a streaming setting.  Online methods can vary in whether their focus is to rapidly detect a single change point (e.g.,~\cite{xie2021sequential}) or to detect a sequence of multiple change points (e.g.,~\cite{liu2013change,li2015scan,chang2018kernel,cheng2020optimal}). 
%In many real world settings, change point methods need to be applied on streaming data, which makes it infeasible to use many offline partitioning methods. Online change point detection methods either focus on detecting a single change point in the quickest possible manner  or use test statistics (e.g., the Maximum Mean Discrepancy (MMD) or Wasserstein distances) on sliding windows to detect multiple change points
%\cite{liu2013change,li2015scan,chang2018kernel,cheng2020optimal}.
What nearly all of these approaches have in common is that they are fundamentally \emph{unsupervised}. This makes it particularly challenging to identify subtle changes, especially in the high-dimensional setting. In such problems we are hoping to find a needle in a haystack, but without knowing what a needle looks like!

Fortunately, in many settings we actually have access to a limited number of expert-labeled change points. This offers a potentially powerful way to provide information about what kind of changes we wish to detect, what kind of changes we may wish to ignore, and what features are most relevant to our task. %A natural question to ask is how can we incorporate available true change point information to improve change point detection?
%
%Consider the example given in Figure \ref{Fig: BeeDance_complt_example} where a 3 dimensional input is undergoing various changes. 
%If we do not consider the true change points, one can argue that the red and blue signals are undergoing a change in the slope, but we  can also say that a non-parametric distribution is generating an oscillatory sequence, and there isn't a change in this distribution.
%We can observe that true changes, shown by vertical dashed lines, are often associated with changes in variance of the yellow signal. 
%
%
In this paper we propose a framework for using labeled change points to learn a \emph{ground metric} for improved online change point detection. We show that by learning a metric that highlights changes of interest, we can both improve change point detection (over unsupervised and non-metric based approaches) and also reveal interpretable maps of which features are most important.
\pagebreak

The main contributions of our work are:
%\vspace{-4mm}
\begin{itemize}[leftmargin=13pt]
\setlength\itemsep{0mm}

    \item We propose SinkDivLM, a novel approach for change point detection that uses available  change points labels to obtain similarity triplets for learning a ground metric for Sinhkorn divergences. These learned Sinkhorn divergences are then used in two-sample tests over sliding windows for change point detection in the online setting.
   % \item We propose a method, which to our knowledge, is the first method that uses   true change points to learn a metric for  change point detection. This involves devising a way to generate similarity triplets from available true change points labels. These similarity triplets are then used to learn a ground metric for Sinkhorn divergence based two-sample tests over sliding windows to detect change points.
    \item The ground metric learned through our approach provides flexibility and can be suitable for different change point detection settings. For example, by incorporating an $\ell_1$-norm penalty, we show how a sparse ground metric can be learned for simultaneous feature selection and change point detection in high-dimensional settings. 
        
    %\item The learned metric can be useful for interpreting what features are important, or how features are related, for detecting change points.
    \item We experiment on simulated as well as real world sequences to show that our proposed method can improve change point detection performance and reveal important features and interactions needed to reliably detect change points in both low- and high-dimensional settings.
    %\item Show that the method provided a way to project sequences in a lower dimensional space through supervision.
\end{itemize}
%Modern sequences are often generated by non-parametric distributions undergoing changes of various kinds, and often, only changes of a certain kind are relevant. 

%We use available change points to learn a ground metric for Sinkhorn Divergences, that are normalized variants of entropic regularized Wasserstein distances. These Sinkhorn divergences, equipped with the learned ground metric, are then repeatedly used in two-sample tests on overlapping sliding windows to detect change points.

\section{Background and related work}

\subsection{Change point detection}
The most widely used algorithms for change point detection are parametric approaches such as Cumulative SUM (CUSUM)  and Generalized Likelihood Ratio (GLR). 
Both of these methods model changes as parametric shifts in the underlying distributions and operate on statistics formed from the log-likelihood ratio between the pre-change and post-change distributions~\cite{basseville1993detection,xie2021sequential}.
%, that is either known in advance, in the case of CUSUM, or is estimated for GLR.  A survey on such methods can be found in \cite{basseville1993detection,xie2021sequential}.
These methods are best suited to  detecting a single change in the  \emph{quickest} time (given some false alarm constraint) in settings where simple parametric models are realistic.

Over the past decade, there has been an increasing focus on \emph{non-parametric} change point detection methods that operate over sliding windows. 
 Integral probability metrics such as the kernel ratio, the maximum mean discrepancy (MMD), and Wasserstein distances are then used in a two-sample test detect change points \cite{liu2013change,li2015scan,cheng2020optimal}. Other kernel  methods such as the Hilbert Schmidt Independence Criterion with $\ell_1$ regularization (HSIC Lasso) have used sliding windows for feature selection in high dimensional change point settings \cite{yamada2013change}. One potential problem for kernel based non-parametric change point detection methods is that it is difficult to tune the bandwidth parameter for the commonly-used radial basis function (RBF) kernel that is used in MMD. This can lead to undesirable performance in some regimes. Wasserstein distances provide an alternate way to conduct two-sample tests without needing to tune a bandwidth parameter. 
Some recent works have also used neural network autoencoders and GANs to learn (unsupervised) representations that can then be used in sliding windows to detect change points \cite{chang2018kernel,deryck2021change}.  

All of these methods detect change points in an unsupervised way.  
There is, however, limited work that incorporates supervision in the form of labeled change points.  Most of this work (see~\cite{li2006supervised,lajugie2014large,aminikhanghahi2017survey}) is restricted to the offline change point setting, as opposed to the online setting that we consider in this paper.
% change point instances \cite{li2006supervised} as labels.
%Most of these methods were proposed more than a decade ago and there isn't much work, at least in our knowledge, that combines recent non-parametric change point detection methods with available information on where the true changes are. 
%In the online setting, the most relevant prior work is an approach that uses pseudo labels for detecting change points through feature selection in high-dimensional settings using the Hilbert Schmidt Independence Criterion with $\ell_1$ regularization (HSIC Lasso) \cite{yamada2013change}. %The main idea in  to construct a pseudo change label sequence at each time instance, where this label sequence has a value of +1 ,and a value of -1 after the instance being tested for a change point. The Hilbert Schmidt Independence Criterion with $\ell_1$ regularization (HSIC Lasso) is then calculated between input sequence features and the constructed pseudo change label sequence.
%While selecting relevant features, the HSIC Lasso criterion is used to detect a change point. HSIC, however, is a statistical independence test and not a two-sample test that can be used to control type 1 and type 2 errors in hypothesis testing settings that are commonly used in change point detection. However, the HSIC Lasso Criterion is mostly restricted to just feature selection in change point detection settings. In many scenarios, all features in an input sequence might be relevant to detect a change point but these features might be weighted in a different manner.

\subsection{Wasserstein distances and Sinkhorn divergences}
Wasserstein distances compute the minimal cost of transporting mass from one distribution to another. Concretely, consider two discrete multivariate distributions $\valpha$ and $\vbeta$ on $\R^d$. We can express these distributions as 
\[
\valpha = \sum_{i = 1}^n a_i \vdelta_{\vx_i} \quad \text{and} \quad  \vbeta =  \sum_{j = 1}^m  b_j \vdelta_{\vy_j}, 
\] 
where $\vdelta_{\vx} $ is the Dirac function at position $\vx \in \R^d$, so that the $\vx_i$ and $\vy_j$ denote the mass locations for the distributions and $a_i, b_i \in \R_+$ are the weights at these mass locations for $\valpha$ and $\vbeta$ respectively.  %$\alpha_i$ and $\beta_i$ lie in $\R$ and are the corresponding weights for each mass location.  
The  ground cost metric  $\mC \in \R^{n \times m}$   represents the transportation cost  between each pair of distribution mass locations. In this work, we consider \textbf{Wasserstein 2} ($\mathcal{W}^2$) distances that use a squared distance ground cost metric, where the $(i,j)^{\text{th}}$ entry of $\mC$ is given by
\[ 
\mC_{i,j}  = \| \vx_i - \vy_j  \|_2^2 . 
\]
As the goal is to minimize the cost of moving mass between two distributions, Wasserstein distances require computing a transport plan $\mP$ that dictates how mass is transported between the distributions.  This is done by solving the following optimization problem:
\begin{align}
    &\mathcal{W}(\valpha, \vbeta ) =  \min_{\mP }   \langle \mC, \mP\rangle, 
   \nonumber \\
    &\text{ subject to } \mP  \in \R_{+}^{n \times m}, 
    \mP^T\mathbbm{1}_n = \vb, \mP \mathbbm{1}_m = \va  \nonumber,
\end{align}
where $\langle \mC , \mP\rangle $ is the Frobenius inner product between the cost matrix $\mC$ and the transport plan $\mP$, $\va$ and $\vb$ contain the mass weights for the distributions $\valpha$ and $\vbeta$, and $\mathbbm{1}_n \in \R^n$ is the vector of all ones. 
%This inner product can be explicitly written as \[ \langle \mC , \mP\rangle = \sum_{i}^n\sum_{j}^n \mP_{i,j}\|\vdelta_i - \vdelta_j \|_2^2,  \] 

Wasserstein distances can be unstable and computationally expensive to compute, requiring $O(n^3\log n)$ computations to evaluate in the case where $n$ and $m$ are of the same order. This makes it difficult to use Wasserstein distances repeatedly in two-sample tests. Additionally, the minimization problem can also be sensitive to slight changes in the input.
%\begin{align}
%    \mathcal{W}_\gamma (\valpha, \vbeta ) =  \min_{\mP }  & \langle \mC , \mP\rangle  - \gamma \mH(\mP) \label{eq: reg OT} \\
%   =\min_{\mP }&  \sum_{i}^n\sum_{j}^n \mP_{i,j}\|\vdelta_i - \vdelta_j \|_2^2 + \gamma\mP_{i.j}(\log\mP_{i,j} - 1)
 %  \nonumber  \\ 
%   &\text{ subject to }  \\
%    &\mP  \in \R_{+}^{n \times m} \nonumber \\ 
%    &\mP^T\mathbbm{1}_n = \vb, \mP\mathbbm{1}^T_n = \va  \nonumber,
%\end{align}
One solution to these problems is to add a regularization term $\mH(\mP)$ to form the entropic regularized Wasserstein distance $\mathcal{W}_\gamma$ \cite{cuturi2013sinkhorn,peyre2019computational}. This is  also known as the Sinkhorn distance and is defined as
\begin{align}
    &\mathcal{W}_\gamma (\valpha, \vbeta ) =  \min_{\mP }   \langle \mC , \mP\rangle  - \gamma \mH(\mP) \label{eq: reg OT} ,\\
   &\text{ subject to }
    \mP  \in \R_{+}^{n \times m} ,
    \mP^T\mathbbm{1}_n = \vb, \mP\mathbbm{1}_m = \va  \nonumber,
\end{align} 
where $\mH(\mP)$ is the entropy of the transport plan matrix $\mP$ and is given by 
\[
\mH(\mP) = \sum_{i=1}^n \sum_{j=1}^m \mP_{i.j}(\log\mP_{i,j} - 1),
\] 
while $\gamma$ is a regularization parameter. This regularization terms makes the minimization problem convex, which makes it less sensitive to changes in input, and can be solved with $O(n^2)$ computations using the Sinkhorn algorithm \cite{cuturi2013sinkhorn}.
Note that the regularized Wasserstein distance is biased as $\mathcal{W}_\gamma^2 (\valpha, \valpha) \neq 0$. An unbiased divergence can be constructed from these regularized Wasserstein distances and is called the \textbf{Sinkhorn divergence}:
%\[ S_{\gamma}(\valpha,\vbeta) =  \mathcal{W}_{\gamma} (\mathbf{a}, \mathbf{b} ) - \frac{1}{2}\mathcal{W}_\gamma (\mathbf{a}, \mathbf{a}) -\frac{1}{2} \mathcal{W}_\gamma (\mathbf{b}, \mathbf{b} )        \]
\begin{equation}
    S_{\gamma}(\valpha,\vbeta) =  \mathcal{W}_{\gamma} (\valpha, \vbeta ) - \frac{1}{2}\mathcal{W}_\gamma (\valpha, \valpha) -\frac{1}{2} \mathcal{W}_\gamma (\vbeta, \vbeta ).
    \label{eq: Sinkhorn Div}
\end{equation}

The regularization parameter $\gamma$ allows Sinkhorn divergences to interpolate between Wasserstein distances and energy distances \cite{feydy2019interpolating,ramdas2017wasserstein}. %Though Sinkhorn divergences also have a $\gamma$ parameter, they are less  sensitive to this parameter, with respect to different regimes within a sequence, as kernel based methods are to the rbf bandwidth parameter \cite{peyre2019computational}.

\subsection{Learning a ground metric for optimal transport}
While the squared distance is a natural choice for the ground cost metric, when it is available, side information can also be used to learn an improved ground metric. This idea was first explored to directly estimate the ground cost given similarity/dissimilarity information for  nearest neighbour classification tasks \cite{cuturi2014ground}. Similarity/dissimilarity information was also used to learn a Mahalanobis ground metric to compare word embeddings through Wasserstein distances in \cite{huang2016supervised}. Unsupervised ground metric learning has been also leveraged to devise subspace robust Wasserstein distances \cite{paty2019subspace} that lead to better Wasserstein distance performance in high dimensional settings. This is done by findind an orthogonal projection of a given rank on the input data such that the Wasserstein distance between samples is maximized. Ground metric learning has also been used to compare entire time series/sequences using order preserving Wasserstein distances \cite{su2019learning}. In such settings, time series labels are used to learn a ground metric.  %Similar sequences in such a setting, however, assume  that they have similar starting points, an assumption that is violated in windowed sub-sequences in change point settings. 
Other applications involving ground metric learning include domain adaptation and label distribution learning 
\cite{zhao2018label,kerdoncuff2021metric}.
%\subsubsection{A toy example}
%begin{figure}[h]
%    \centering
%    \includegraphics[width = 0.5\textwidth]{figures/toy_example_Grnd_Metric.pdf}
%    \caption{Toy example - Three different discrete distribution $\valpha$, $\vbeta$ and $\boldsymbol{\phi}$. Suppose we have some supervised information requiring  $S_{\gamma}(\valpha,\vbeta)$ %( or $\mathcal{W}_{\gamma} (\valpha, \vbeta )$)%
%    to be large but $S_{\gamma}(\valpha,\vphi)$ %(or $\mathcal{W}_{\gamma} (\valpha, \vphi )$)%
%    to be small. How can this be done? One simple way is to eliminate the $\vy$ axis for all mass locations $\vdelta$ by performing a linear transformation $\mL\vdelta$ , where $\mL =$ $\begin{pmatrix}
%  1 & 0\\ 
%  0 & 0
%\end{pmatrix}$  . Such a transformation would lead to $S_{\gamma}(\mL\valpha, \mL \vphi)  = 0$, as $\mL\valpha = \mL \vphi $.}
%    \label{Fig:Toy_example_Grnd_Metric}
%\end{figure}

%Consider the case where we are given three distributions $\valpha,\vbeta$ and $\vphi$ as shown in Figure \ref{Fig:Toy_example_Grnd_Metric}. We are asked to measure divergences between distributions but with a caveat;  We are required to obtain a smaller divergence between distributions of type $\valpha$ and type $\vphi$ than the divergence between distributions of type $\valpha$ and  type $\vbeta$. One way to satisfy this requirement is to apply a linear transform $\mL = $ $\begin{pmatrix}
% 1 & 0\\ 
%  0 & 0
%\end{pmatrix}$  on the mass locations $\vdelta$. 
%Such a transform would lead to $S_{\gamma}(\mL\valpha, \mL \vphi)  = 0$. 

\subsection{Sinkhorn divergence with learned ground metric}

%We can now formally equip Sinkhorn divergences with such transformations.
A learned ground metric can be readily incorporated into the calculation of the Sinkhorn divergence. Suppose that we have learned a Mahalanobis metric parameterized by an inverse covariance matrix $\mM$ with rank $r$, and consider the factorization $\mM = \mL^T \mL$, where $\mL$ is an $r \times d$ matrix.  For mass weight locations $\vx_i, \vy_j  \in \R^d$, 
we can express the ground cost matrix in terms of this Mahalanobis distance as the matrix $\mC_{\mL}$ with  $(i,j)^{\text{th}}$ element given by
\[ 
{\mC_{\mL}}_{i,j}  = \| \mL ( \vx_i - \vy_j)  \|_2^2 . 
\]
$\mC_{\mL}$ can then be used to compute the Sinhkorn distance: 
%\begin{align}
%    \mathcal{W}_{\mL,\gamma} (\valpha, \vbeta ) =  \min_{\mP }   \langle \mC_{\mL} ,
%    \mP\rangle  - \gamma \mH(\mP) \label{eq: reg OT} \\
%= \min_{\mP } &  \sum_{i}^n\sum_{j}^n \mP_{i,j}\|\mL(\vdelta_i - \vdelta_j)\|_2^2 + %\gamma\mP_{i,j}(\log\mP_{i,j} - 1)
%   \nonumber  \\ \text{ subject to }  
%    &\mP  \in \R_{+}^{n \times m} \nonumber \\ 
%    &\mP^T\mathbbm{1}_n = \vb, \mP\mathbbm{1}^T_n = \va  \nonumber,
%\end{align} 
\begin{align}
    &\mathcal{W}_{\mL,\gamma} (\valpha, \vbeta ) =  \min_{\mP }   \langle \mC_{\mL} ,
    \mP\rangle  - \gamma \mH(\mP) \label{eq: reg OT Sink}, \\
   &\text{ subject to }
    \mP  \in \R_{+}^{n \times m},
    \mP^T\mathbbm{1}_n = \vb, \mP\mathbbm{1}_m = \va  \nonumber.
\end{align}
As before, these paramaterized Sinkhorn distances can be used to obtain the parameterized Sinkhorn divergence:
\[
S_{\mL,\gamma }(\valpha,\vbeta) =  \mathcal{W}_{\mL, \gamma} (\valpha, \vbeta ) - \frac{1}{2}\mathcal{W}_{\mL,\gamma} (\valpha, \valpha) -\frac{1}{2} \mathcal{W}_{\mL, \gamma} (\vbeta, \vbeta ) .
\]
%In literature, the parameterized Wasserstein distance $ \mathcal{W}_{\mL,\gamma} (\boldsymbol{\mathcal{\alpha}}, \vbeta )$ is sometimes denoted by $\mathcal{W}_{\gamma} ( f\# \va, f \#  \vb )$, where $ f \# \va$ denotes sample $\valpha$  from distribution $\va$ being pushforwarded by the transformation $ f $ i.e, $f \# \va = f(\valpha) = \mL^T\valpha$, where $\valpha \sim \va$ and $f$ is the linear transform $\mL$. Transformations $f$ could be generalized to other non-linear transformations too, but we are considering linear transformations - initially at least.
\section{Proposed method}

\subsection{Sinkhorn divergence on sequences }
Before considering how supervised change detection can be performed by combining Sinkhorn divergence with a learned ground metric, we first clarify how Sinkhorn divergences can be applied to compare two sequences.
Consider two sequences $\mX \in \R^{t  \times d}$ and $\mY \in \R^{t \times d}$, where  $t$ is the length of the two sequences (which for simplicity we assume to be equal) and $d$ is the dimension of each sample in the two sequences. We can construct the empirical distribution of $\mX$ and $\mY$ via
\[ 
\sum_{i=1}^{t}  \frac{1}{t}\vdelta_{\vx_i} \quad \text{and} \quad \sum_{i=1}^{t}  \frac{1}{t}\vdelta_{\vy_i}
\]
These empirical distributions take uniform weights of $\frac{1}{t}$ at the mass locations. While there are many other ways to represent the sequences as discrete distributions, this scheme is often used because of it simplicity.
The parameterized Wasserstein distance between the two sequences\footnote{Note that we will slightly abuse notation in writing $\mathcal{W}_{\mL,\gamma} (\mX, \mY )$ by letting $\mX$ and $\mY$ denote both the empirical distribution of the sequences as well as the sequences themselves.}  can be computed as:
\begin{align}
    \mathcal{W}_{\mL,\gamma} (\mX, \mY ) 
=  \min_{\mP } &  \sum_{i,j=1}^{t} \mP_{i,j}\|\mL(\vx_i - \vy_j) \|_2^2 - \gamma \mH(\mP) 
      \label{eq: OT plan parameterized} \\ \text{ subject to }  
    &\mP  \in \R_{+}^{t \times t}, 
    \mP^T\mathbbm{1}_{t} = \mathbbm{1}_{t} , \mP\mathbbm{1}_{t} = \mathbbm{1}_{t} \nonumber.
\end{align} 

\begin{figure}[t!]
    \centering
    \includegraphics[width = 0.8\textwidth]{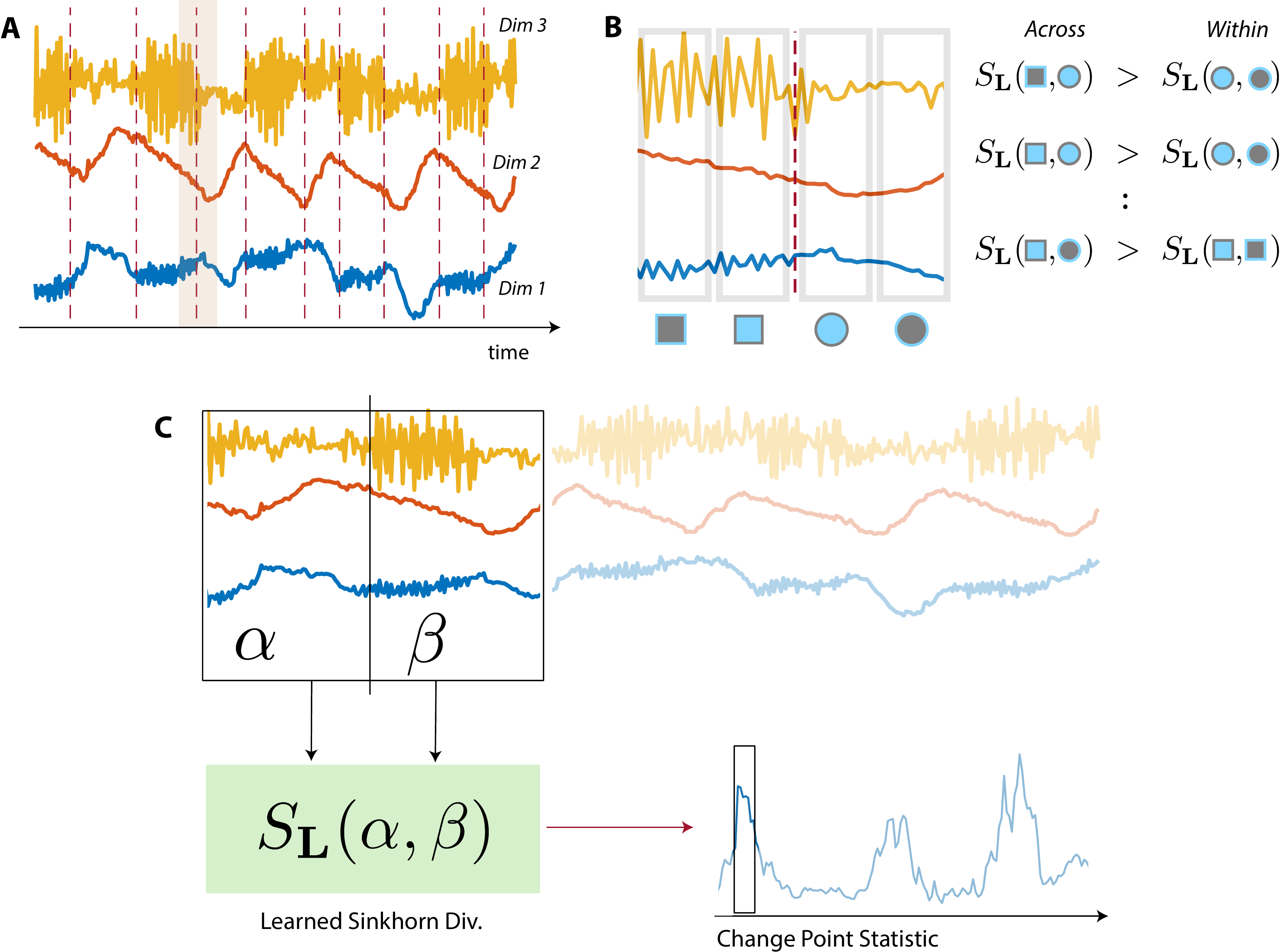}
    %\p3ion{For the beedance dataset we can see that the true change instances are mostly associated with variance changes in the yellow signal. Unsupervised change point detection methods detect spurious changes leading to poor performance. We investigate how to use some of these true change instances to improve change point detection performance}
   % \vspace{-2mm}
    \caption{In (A-B), we show how labeled change point instances (red vertical lines) are used to obtain similarity triplets from pre-change (square sub-sequences) and post-change windows (circled sub-sequences) as shown in B. These triplets are then used to learn a linear transformation $\mL$ that ensures samples across the change points are far apart. (C) After learning $\mL$, we use a two-sample test in a sliding window to perform online change point detection.}
    \label{Fig: BeeDance_complt_example}
\end{figure}

\subsection{Generating similarity triplets from change points}

%\begin{figure}[t!]
%    \centering
%    \includegraphics[width = 0.5\textwidth]{figures/true_cp_in_multivar.pdf}
%    \caption{A 2 dimensional signal with a provided true change point at instance $i$. The true change point is associated with a change in only one - black - dimension of the signal. The true change point provides a set of pairs of similar ($\{(X_1^p,X_2^p),(X_1^f,X_2^f)\}$) and dissimilar sub-sequences ($\{(X_2^p,X_1^f),(X_2^f,X_1^p)\}$) that can be used to learn a ground metric for Sinkhorn divergences. Similarly, there could be cases where the true change is caused by a weighted linear combination of dimensions, rather that only a subset of dimensions }
%    \label{Fig:True CP segments}
%\end{figure}

As shown in Figure \ref{Fig: BeeDance_complt_example}, a true change point can be used to generate similar and dissimilar pairs of sub-sequences. In \ref{Fig: BeeDance_complt_example}(B),  we obtain two sub-sequences represented by the grey and blue squares. Mathematically, we refer to these as $\mX_1^p$ and $\mX_2^p$ respectively. Similarly we can obtain two sub-sequences after the change point that are shown in blue and grey circles. We  refer to these as $\mX_1^f$ and $\mX_2^f$.
%Though there are visible changes in both dimensions of the bi-variate sequence, only one of these dimensions is responsible for the true change given in red. There could be scenarios, where a change is caused by a linear combination of some sub-set of sequence dimensions.&
Sub-sequences on the same side of the change should be similar, whereas the sub-sequences on the opposite side of the change points should be dissimilar. This can be captured mathematically via the Sinkhorn divergence via a constraint that, for example, $S_{\mL, \gamma}(\mX_1^p,\mX_2^p)$ should be smaller than $S_{\mL,\gamma}(\mX_1^p,\mX_2^f)$. Such constraints can be represented as triplets $(\mX_i,\mX^s_i,\mX^d_i)$, where $\mX^s_i$ represents a sequence that that is nearer to $\mX_i$ than $\mX^d_i$.  From each labelled change point we can construct such triplets. 

 \subsection{Learning a ground metric for change detection}
 Our goal is to leverage the triplets generated from the labelled change points to learn a ground metric such that the Sinkhorn divergence $S_{\mL, \gamma}$ does a better job of highlighting change points. The metric learning community has been considering similar problems in a range of works works \cite{weinberger2009distance,xing2002distance}. When comparing distributions, the Wasserstein distance, or its variants such as the Sinkhorn divergence, can capture differences in geometry of samples in a distribution, helping detect samples with dissimilar distributions. Equipping Sinkhorn divergences with a learned ground metric can further improve improve this ability by transforming the data in a way that highlights dissimilarities (and de-emphasizes similarities). This can be done by using the triplet loss
\begin{equation}
    l(\mL) = \sum_{i \in \text{Triplets}}  %\mathcal{S}_{\mL,\gamma} (\mX_i, \mX_{i}^s ) + %
    \left[c - (\mathcal{S}_{\mL,\gamma} (\mX_i, \mX_{i}^d) - \mathcal{S}_{\mL,\gamma} \left(\mX_i, \mX_{i}^s\right) )\right   ]^+,
    \label{eq: loss func}
\end{equation}   
% \begin{align*}
%    &\min_{\mL,\xi_i}\sum_{i \in \text{Sim}} S_{\mL \gamma}(\mX_i , \mX^s_i) + \sum_i\xi_i \\
%~~&\text{subject to,}~ \\
% &S_{\mL,\gamma}(\mX , \mX_i^d) - S_{\mL,\gamma}(\mX , \mX_{d}^i) \geq \tau - \xi_i  ~~ \forall i, \\
% &\mM \succcurlyeq 0.
%\end{align*}
where $c$ is the triplet margin, $[ \cdot ]^+$ is the hinge loss, and $S_{\mL,\gamma}$ is the parameterized Sinkhorn divergence from \eqref{eq: Sinkhorn Div}.
The gradient of the parameterized Sinkhorn divergence between two sequences $\mX$ and $\mY$ with respect to $\mL$ can be computed as
\begin{align*}
  \frac{\partial \mathcal{S}_{\mL}(\mX,\mY)}{\partial \mL} = 2\mL\sum_{i,j=1}^t \mP^*_{i,j}(\vx_i - \vy_j)(\vx_i - \vy_j)^T,
\end{align*}
where $\mP^*$ is the optimal transport plan computed by solving \eqref{eq: OT plan parameterized}.
The gradient of the triplet loss function in \eqref{eq: loss func} is
\begin{equation}
    \frac{\partial l(\mL)}{\partial \mL} =%\sum_{i \in \text{All trip pairs}} 2\mL\sum_{k=1}^n\sum_{j=1}^n \mP^*_{i_k,i^s_j}(\vx_{i_k} - \vx_{i_j}^s)(\vx_{i_k} - \vx_{i_j}^s)^T + \\
  \sum_{v \in \text{Viol}}2\mL \sum_{i,j=1}^t \mP^*_{v_i,v^s_j}(\vx_{v_i} - \vx_{v_j}^s)(\vx_{v_i} - \vx_{v_j}^s)^T  - 2\mL\sum_{i,j=1}^t \mP^*_{v_i,v^d_j}(\vx_{v_i} - \vx_{v_j}^d)(\vx_{v_i} - \vx_{v_j}^d)^T,
  \label{eq:loss gradient L}
\end{equation}
where %$i$ is the index for all similarity triplets%
$v$ is the index for similarity triplets that violate the hinge loss constraint in \eqref{eq: loss func}, $\mP^*_{v,v^s}$ is the transport plan between $\mX_v$  and its similar pair $\mX_v^s$, and $\mP^*_{v,v^d}$ is the transport plan between  $\mX_v$  and its dissimilar pair $\mX_v^d$. Algorithm \ref{Alg: 1} shows how this gradient can be used to learn the transformation $\mL$. Algorithm \ref{Alg: 2} shows how this learned transformation can be used for change detection.
\begin{algorithm}[t]
\begin{algorithmic}
\caption{Learn transform $\mL$ for ground metric using similarity triplets from true change points}
\label{Alg: 1}
\State \textbf{Inputs:} Set of triplets $(\mX_i,\mX_i^s,\mX_i^d)$ from true change points, Sinkhorn regularization parameter $\gamma$, Gradient descent rate $\mu$, Triplet loss constant $c$
\vspace{1mm}
%$\mathcal{X}_S^l, \mathcal{X}_D^l$ : Sets of similar/dissimilar pairs from labels
\State \textbf{Output:} Trained $\mL$
\vspace{1mm}
\State \textbf{Initialize:} $\mL_0$
\vspace{1mm}
    \For{$t = 1$ to number of iterations}
    \State Identify triplet indices $v$ that violate the hinge constraint %in \eqref{eq: loss func}
    \State Compute transport plan between similar pairs $\mP_{v,v^s}^*$ by solving  $\mathcal{S}_{\mL_{t-1}}(\mX_v,\mX_v^s)$  $\forall$  $v$  %using \eqref{eq: OT plan parameterized}
     \State Compute transport plan between dissimilar pairs $\mP_{v,v^d}^*$ by solving  $\mathcal{S}_{\mL_{t-1}}(\mX_v,\mX^d_v)$ $\forall$  $v$  %using \eqref{eq: OT plan parameterized}
     \State Use computed transport plans $\mP_{v,v^d}^*$ and $\mP_{v,v^s}^*$ to form gradient $\frac{\partial l (\mL)}{\partial\mL}$ %using \eqref{eq:loss gradient L}
    \State $\mL_t = \mL_{t-1} - \mu\frac{\partial l (\mL)}{\partial\mL}$ 
    \EndFor
\end{algorithmic}
\end{algorithm}

\begin{algorithm}
\begin{algorithmic}[b]
\caption{Using Sinkhorn divergence with learned metric for change detection}
\label{Alg: 2}
\State \textbf{Inputs:} Sequence $\mX$, Window length $w$, Change threshold $\tau$, Learned $\mL$
\vspace{1mm}
%$\mathcal{X}_S^l, \mathcal{X}_D^l$ : Sets of similar/dissimilar pairs from labels
\State \textbf{Output:} Detected changes
\vspace{1mm}
    \For{$n = 1$ to length of sequence $\mX$}
     \State Form consecutive windows $\mX_p^n,\mX_f^n$ at index $n$
     \State $m_n = S_{\mL,\gamma}(\mX_p^n,\mX_f^n)$ using \eqref{eq: Sinkhorn Div}
     \If{$m_n > \tau$}   
        \State Add $n$ to change points
        \EndIf
    \EndFor
\end{algorithmic}
\end{algorithm}

\subsection{Learning a sparse ground metric}

Additional regularization terms can be used in conjunction with the triplet loss to learn a ground metric that is suitable for different change detection settings. For example, adding a regularizing with an $\ell_1$ or a mixed norm loss has been used for learning a sparse metric \cite{ying2009sparse}. We can use the same idea for learning a sparse ground metric by considering, for example,
\[\min_{\mL}  %\mathcal{S}_{\mL,\gamma} (\mX_i, \mX_{i}^s ) + %
    l(\mL) + \lambda \|\mL\|_1.
    \label{eq: loss reg}
\]
Such an approach aims to learn a metric that depends on only a sparse subset of the original features in $\R^d$.

\subsection{Learned metrics can improve two-sample tests using Sinhkorn divergences }
Sample complexity results for Sinkhorn distances, $\mathcal{W}_{\gamma}$, were given in \cite{genevay2019sample}. A straightforward extension of these results can be obtained for  Sinkhorn divergences.
%The authors worked with the dual potentials (or dual Lagrangian coefficients), associated with the dual formulation of  \ref{eq: reg OT} and showed these potentials lie in a RKHS associated with Sobolov Kernels. A straightforward extension of these results can be obtained for  Sinkhorn divergences.
\pagebreak
\begin{proposition}
\label{thm: smp_cmpx}
If $n$ samples are used to estimate the empirical distributions $\widehat{\valpha}_n \sim \valpha$  and $\widehat{\vbeta}_n \sim \vbeta$ on $\R^d$, then the deviation of the Sinkhorn divergence between these empirical distributions from the true distributions is bounded with probability $1-\delta$:
% \[        \]
\begin{equation}
        |S_{\gamma}(\valpha,\vbeta) - S_{\gamma}(\widehat{\valpha}_n,\widehat{\vbeta}_n) | \le 6B\frac{\lambda K}{\sqrt{n}} +C\sqrt{\frac{8\log(\frac{2}{\delta})}{n}},
    \label{eq: power}
\end{equation}
where $\lambda = O(\max(1,\frac{1}{\gamma^{d/2}})$, $B$ and $C$ are constants, and $K$ is the kernel associated with the dual Sinkhorn potentials.
\end{proposition}
%More details for these constants are provided in the Appendix section.
These sample complexity results can be used to obtain deviation bounds for the Sinkhorn divergence under the null distribution. 
\begin{corollary}
The Sinkhorn divergence between two $n$ samples $\valpha_n^1,\valpha_n^2 \sim \valpha$, is bounded by
\begin{equation}
    |S_{\gamma}(\widehat{\valpha}_n^1,\widehat{\valpha}_n^2) | \leq  6B\frac{\lambda K}{\sqrt{n}} +C\sqrt{\frac{8\log(\frac{2}{\delta})}{n}}. 
    \label{eq: Null}
\end{equation}  
%where again $\lambda = O(\max(1,\frac{1}{\gamma^{d/2}})$, $B$ and $C$ are constants, and $K$ is the kernel associated with the dual Sinkhorn potentials.
\end{corollary}
As the metric learning loss in \eqref{eq: loss func} contains similar pairs from the same distribution, an ideally learned metric would ensure that $S_{\mL,\gamma}(\valpha,\valpha) =0$, while $S_{\mL,\gamma}(\valpha,\vbeta) \geq c$ for dissimilar pairs, where $c$ is the triplet loss margin. In other words, the Sinkhorn divergence between similar pairs from the same distribution would be 0 and the Sinkhorn divergence between dissimilar pairs would be greater than the margin $c$. If this margin $c$ is set such that  $c > S_\gamma(\valpha,\vbeta)$, i.e., greater than the Sinkhorn divergence without a learned metric, then it is likely that $S_{\mL,\gamma}(\valpha,\vbeta)  > S_\gamma(\valpha,\vbeta)$. From~\eqref{eq: power}, this will likely ensure that $S_{\mL,\gamma}(\widehat{\valpha}_n,\widehat{\vbeta}_n)  > S_\gamma(\widehat{\valpha}_n,\widehat{\vbeta}_n)$, resulting in a Sinkhorn divergence with increased testing power.

Under the case where case, where both samples come from the same distribution, the results in \eqref{eq: Null} show that
when the regularization parameter $\gamma$ is small, a large dimension of the input distributions can lead to a large Sinkhorn divergence, which would result in false change points.
Learning a ground metric learning allows us to enforce a structure on the distribution that improves sample complexity results. This can be done by projecting the distribution into low-dimensional subspace, as explored in \cite{paty2019subspace,wang2021two}.  Learning either a  low-dimensional projection $\mL$ %(which would lead to a low-rank Mahalanobis ground metric parameterized by $\mL^T\mL$) 
or a sparse projection $\mL$  reduces the effective dimension of the data distribution, leading to improved change by detecting fewer false change points.

\section{Experiments}

\subsection{Evaluation metrics}
We use area under the curve (AUC) to report the performance of our change point detection method as it is used by other work papers to evaluate performance \cite{chang2018kernel,cheng2020optimal}.  
Similar to the evaluation methods used in \cite{chang2018kernel}, a change point is detected correctly if it is detected at the true change point location. The AUC metric covers the true positive and false positive rates at different detection thresholds and is thus a suitable metric to evaluate change point detection performance.

\subsection{Synthetic datasets}

\vspace{-1mm}
\paragraph{Switching variance.} We simulate the AR process
\[ 
x_1(t) = 0.6x_1(t-1) -0.5x_1(t-2) + \epsilon_t ,
\]
where $\epsilon_t \sim \mathcal{N}(0,\sigma^2)$ and $\sigma$ switches between $\sigma = 1$ and  $\sigma = 5$ every 100 time steps. We also generate a noise vector $ \left[ x_2(t),\ldots, x_{50}(t) \right]  \sim \mathcal{N}(\mathbf{0}_{49},\mI_{49})) $. We concatenate $x_1 (t)$ with  $\left[x_2(t),\ldots,x_{50}(t)\right] $ to obtain a 50 dimensional vector $\left[x_1(t), \ldots, x_{50}(t)\right] $  where changes are only happening in the first dimension.

\vspace{-2mm}
\paragraph{Switching Gaussian mixtures.} Two 100 dimensional Gaussian mixture distributions
$\valpha = \mathcal{N}(\mathbf{0},\mI) + \mathcal{N}(\mathbf{1},\mathbf{\Sigma}_0) $ and $\vbeta = \mathcal{N}(\mathbf{0},\mI) + \mathcal{N}(\mathbf{1.5},\mathbf{\Sigma}_1)$ were used to simulate a sequence where $\valpha$ and $\vbeta$ switched every 100 samples. $\mathbf{\Sigma}_0$ and $\mathbf{\Sigma}_1$ are diagonal covariance matrices where the first 3 entries on the diagonal are 3 and 5 respectively, while the rest of the diagonal entries are 1. A training sequence consisting of 25 changes, of which 80 percent used for training and 20 percent were used for validation, was used to train the ground metric. A separate testing sequence consisting of 25 changes was used to evaluate performance.

\vspace{-2mm}
\paragraph{Switching frequency mixture.} A two dimensional sequence where the first dimension switches between $\sin(2\pi(0.1)t)+\sin(2\pi(0.5)t)+\sin(2\pi(0.3)t)$ and $\sin(2\pi(0.1)t)+\sin(2\pi(0.5)t)+\sin(2\pi(0.35)t)$ every $t=100$. The second dimension switches between  $\sin(2\pi t)+\sin(2\pi(1.5)t)+\sin(2\pi(1.70)t)$ and $\sin(2\pi t)+\sin(2\pi(1.5)t)+\sin(2\pi(0.35)t)$ every $t=100$. These sequences are generated such that there are 10 samples per second. Both these dimension have $\mathcal{N}(0,0.1)$ noise added. 15 changes points are used to train the ground metric, 4 are used for validation. A different sequence consisting of 19 changes points is used as the test set.

\vspace{-2mm}
\paragraph{Switching frequency mixture with slopes.} This includes the frequency switching dataset concatenated with 48 additional dimensions whose slopes change every 1000 samples, resulting in a 50 dimensional dataset. True change point are only labeled at instances where the frequencies in the first two dimensions change. For 24 of the  additional dimensions the slope changes from a gradient of -0.06 to 0.06,  and for the other 24 dimensions the slopes change from 0.06 to -0.06. All of the slope dimensions have $\mathcal{N}(0,0.0001)$ noise added. 

\subsection{Real world datasets}
\vspace{-1mm}
\paragraph{Bee Dance.}
Bees are tracked using videos to obtain three dimensional sequences, where the first two dimensions represent the x,y coordinates for bee location,  while the third dimension shows the bee heading angle. Instances where the bee waggle dance changes from one stage to the other are labelled as change points.  The dataset consists of 6 sequences. %Figure 1 \ref{Fig: BeeDance_complt_example} shows a snapshot from one of these s example of the Bee Dance sequence. %
We used two sequences for training and validation, while the rest of the sequences are used as test datasets. In total there are 15 change points that are used for training, of which 12 are used for training and 3 for validation. 

\vspace{-2mm}
\paragraph{HASC (Activity Detection).}
 HASC-2011 and HASC-2016 datasets consists of people performing performing different activities such as walking, running, skipping, staying. An accelerometer provides a three dimensional sequence where changes are labeled when there is a change in activity. A single sequence from HASC-2016 was used for true labels and training the ground metric. This sequence provided 15 true change points of which 80 percent are used for training and 20 percent for validation. The rest of the  89 sequence datasets in the HASC-2016 are used as test datasets. A single sequence dataset, the same used by \cite{chang2018kernel}, from HASC-2011 is also used as a test dataset.
 
 \vspace{-2mm}
 \paragraph{Yahoo.} 15 sequences containing change points that indicate a change in different metrics, such as CPU usage, memory usage, etc. All sequences are one dimensional. We use 3 change points from one of the 15 sequences to train our metric and use the rest of the sequences for evaluation.
 
 \vspace{-2mm}
 \paragraph{ECG.}  A single dimensional sequence containing change point labels at Ischemia, or abnormal heartbeat, instances. We split the dataset for training and testing, and use 21 change points for training and validation, in an 80-20 split.

\vspace{-2mm}
\paragraph{Mouse Sleep Stage.} This dataset consists of the spiking activity of 42 neurons detected from multi-electrode arrays implanted in the hippocampus of a mouse as it moves through different arousal and sleep stages \cite{azabou2021view,hengen2016neuronal,ma2019cortical} (REm, nREM, and wake) over 12 hours.
 A sub-sequence that contained 14 change points (between REM and nREM) were used to train and validate the learned ground metric. A different sub-sequence consisting of 28 change points was used as a test dataset.

\subsection{Baselines}
We compare the performance of our method (SinkDivML) with different methods that we classify into two categories: those that require access to true change points for learning a model and those that do not.

\vspace{-2mm}
\paragraph{Baselines not requiring access to true change points.}
These baselines include Sinkhorn divergence without learning a metric (SinkDiv), M-stats \cite{li2015scan}, and HSIC \cite{yamada2013change}. SinkDiv and M-stats do not involve any model learning whereas HSIC learns a model using pseudo labels only.

\vspace{-2mm}
\paragraph{Baselines requiring access to true change points.}
These include generative neural network based kernel change point (KLCPD), autoencoder based methods in time domain TIRE\textsubscript{T}, and frequency domain TIRE\textsubscript{F}\cite{deryck2021change}. Though these models are trained in an unsupervised manner, they need access to true change labels to tune and validate the learned model. We use the same sequence datasets to train and validate these models that we use to train and validate our method. We also use a supervised version of HSIC (sHSIC) for experiments that involve feature selection in high dimensional change settings. Rather than using pseudo labels, which HSIC uses, we use true change point labels for feature selection.
\begin{table*}[t]
  \caption{{\em AUC for different change point methods}}
  \centering
  \begin{tabular}{lccccccc}
    \toprule
    Model & Swch GMM & Swch Freq & Bee Dance &  HASC (2011) & HASC (2016) & Yahoo & ECG \\
    \midrule
    HSIC & 0.493 & 0.426 & 0.543 & 0.603 & 0.591 & - & -\\
    M-stats &  0.947 & 0.437 &  0.494  & 0.605 & 0.751 & 0.737 & 0.844  \\
    TIRE\textsubscript{T} & 0.501  &  0.551  &  0.539 & 0.659 & 0.643  & 0.865  & 0.747  \\
     TIRE\textsubscript{F} & 0.677    &  0.647  &  0.556 & 0.725 & 0.712 & 0.871 & 0.900\\
    KLCPD & 0.802 & 0.709 & 0.632 & 0.663 & 0.742 & 0.932  & 0.810\\
    SinkDiv & 0.778 & 0.481 &  0.556    & 0.757 & 0.717 & 0.942 & 0.900\\
    \textbf{SinkDivLM} &0.974 & 0.843 & 0.682 & 0.803 & 0.759 & 0.946 & 0.899  \\
    \bottomrule
    \label{Table: MG result}
  \end{tabular}
\end{table*}

\subsection{Window size and projection dimension settings}
As suggested by the results in the previous section, the empirical Sinkhorn divergence will be closer to the true Sinkhorn divergence as the number of samples increase. Thus a larger sliding window size would generally be better, but would result in increased computation time. The window size is also dataset dependent, with the major constraint being that the windows should not be so large as to cover segments spanning multiple change points. However, it should also be large enough to capture all the aspects of the sequence. For example in the frequency mixture dataset, the window should be long enough to capture at least one complete period of the sinusoids.   The learned projection $\mL$ should ideally project data to a lower dimension to reduce error between true and empirical divergences. However, a larger projection dimension might be needed to actually learn the model. Validation loss can be used to tune both  the projection dimension $r$ and regularization parameter $\gamma$.

\section{Results}
Table \ref{Table: MG result} compares the performance for different change point detection methods on datasets that are either low dimensional, i.e. datasets which  have a maximum dimension of 3, or datasets that involve changes in all dimensions, such as the switching GMM dataset.  As these datasets  are either low dimensional or involve changes in all dimensions, the ground metric is learned without $\ell_1$ regularization. Barring the  the ECG dataset, SinkDivLM performs the best on all datasets. The Bee Dance dataset and HASC 2011 datasets are particularly challenging for other change point detection methods as seen in \cite{chang2018kernel}. For the Bee Dance dataset, SinkDivLM improves performance by 8\% over KLCPD, which is the next best method. On HASC 2011, SinkDivLM leads to a 21\% improvement over KLCPD and a 6 \% improvement over SinkDiv which provided the second best results. This shows that kernel methods that require tuning the bandwidth parameter, such as KLCPD and M-stats, do much worse on HASC 2011 than methods, such as SinkDiv and SinkDivLM, that are based on Optimal Transport.
For HASC 2016 it is easier to detect change points in most of the  90 sequences in the dataset. This leads to  larger scores for most of the baselines, leading to a relatively smaller improvement for SinkDivLM.   The Yahoo and ECG datasets involve abrupt instantaneous changes (and thus involve a very small window size of size 2 and 3 respectively). Our method assumes that changes persist for some time to learn a metric, and these datasets involve transient changes leaving very small sub-sequence windows to learn a metric. For this reason, the performance gain is not prominent on these datasets. Additionally the performance is already relatively strong for other methods on these datasets, leaving relatively till room for improvement. Results for HSIC were not available for one dimensional datasets such as Yahoo and ECG.

When a ground metric is correctly  learned, the Sinkhorn divergence between dissimilar samples is larger than the provided margin constant $c$ in \eqref{eq: loss func}. This results in the change point statistics between  dissimilar sub-sequences being larger than the change point statistics between similar sub-sequences, which makes it easier to set a threshold that correctly detects true  change points with out detecting many false change points. Figure \ref{Fig: GMM change seq} shows how  a learned  metric leads to a larger change statistic between dissimilar sequences on the switching GMM dataset. This leads to improved change point detection performance in Table \ref{Table: MG result}.

Figures \ref{Fig: GMM error vs window} and \ref{Fig: GMM error vs noise} show the relationship between type 1 and type 2 errors between samples from the two 100 dimensional Gaussian Mixture Models, that are used in the GMM switching dataset, at different noise and window sizes. Solid lines represent the Sinkhorn divergence with a metric that is learned without added noise using 10 samples from each distribution, while dashed lines represent Sinkhorn divergence without a learned ground metric. Samples for the null hypothesis were generated using $\valpha = \mathcal{N}(\mathbf{0},\mI) + \mathcal{N}(\mathbf{1},\mathbf{\Sigma}_0) $, while samples for the alternate hypothesis were generated using $\vbeta = \mathcal{N}(\mathbf{0},\mI) + \mathcal{N}(\mathbf{1.5},\mathbf{\Sigma}_1)$. For \ref{Fig: GMM error vs window}, a noise of $\mathcal{N}(\mathbf{0},2\mI)$ was added. In Figure \ref{Fig: GMM error vs noise} 10 samples from each distribution were used.

\begin{figure*}[t]%
\vspace{-4mm}
    \centering
    \begin{subfigure}[Detected changes]{
        \centering
        \label{Fig: GMM change seq}%
        \includegraphics[width=0.310\textwidth]{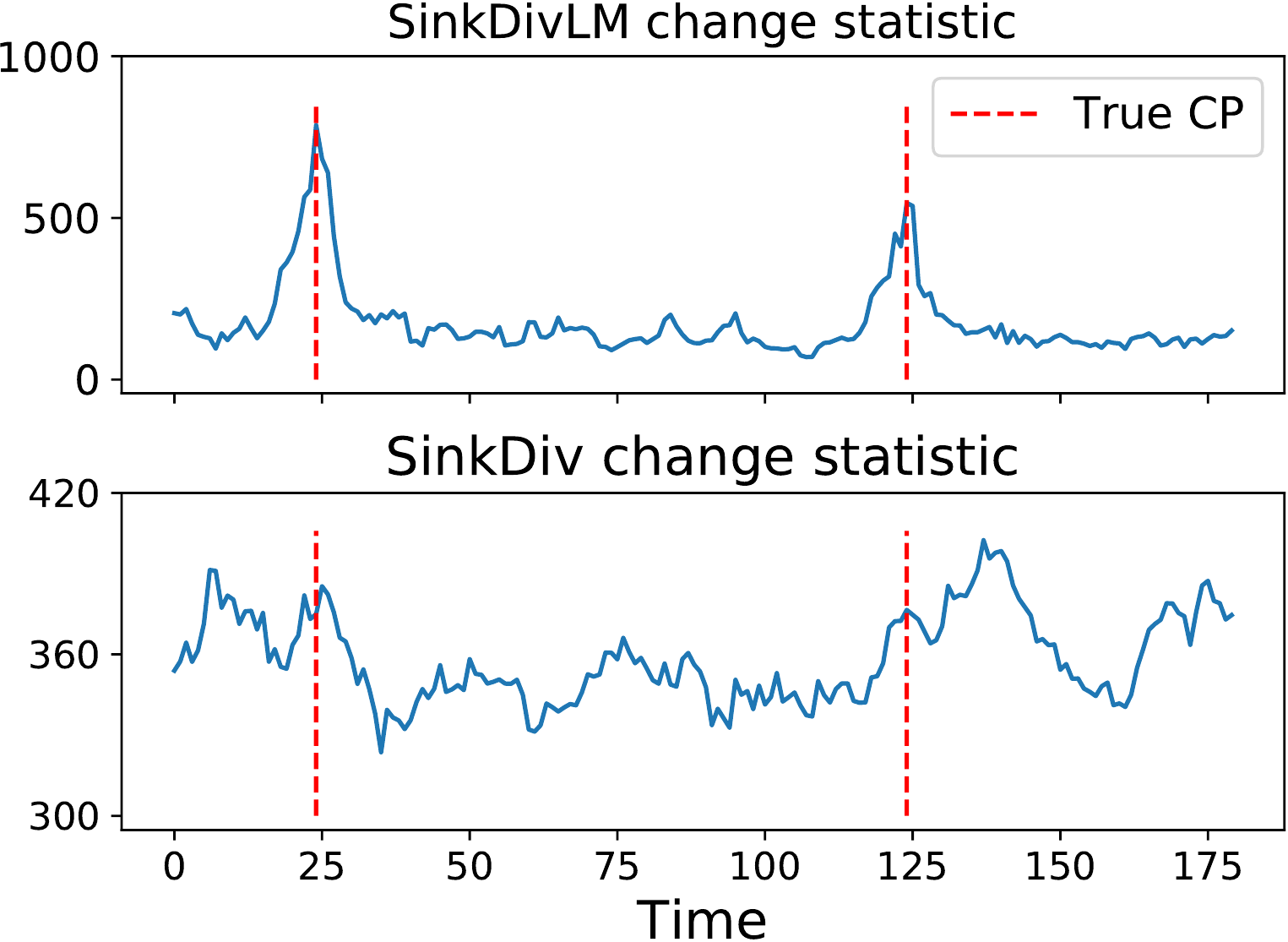}
    }\end{subfigure}
    \begin{subfigure}[Errors across window sizes ]{
        \centering
        \label{Fig: GMM error vs window}%
        \includegraphics[width=0.310\textwidth]{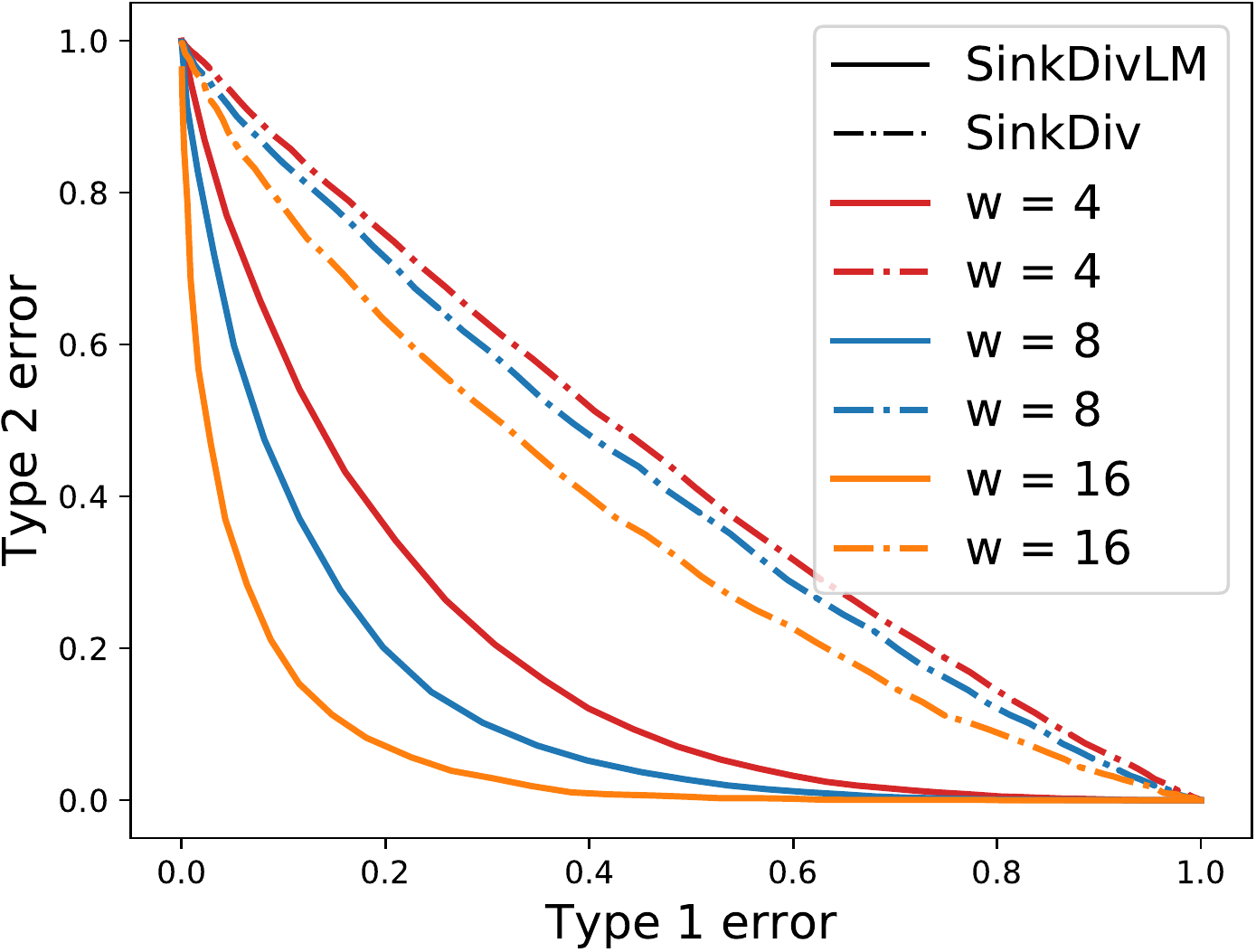}
    }\end{subfigure}
    \begin{subfigure}[Errors across noise levels]
     {   \centering
         \label{Fig: GMM error vs noise}%
         \includegraphics[width=0.310\textwidth]{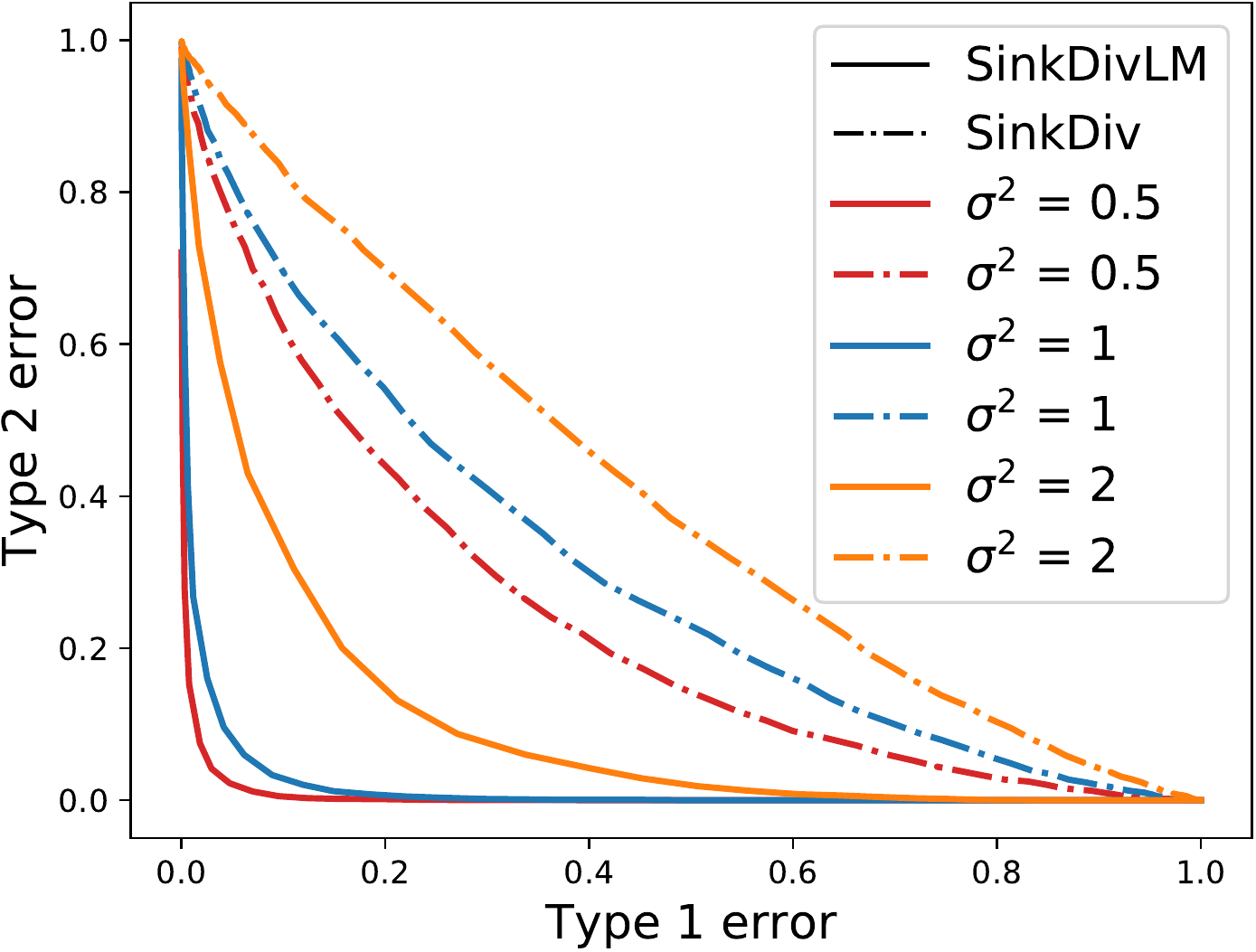}
    }\end{subfigure}
    \vspace{-2mm}
    \caption{{\em Results for the switching GMM dataset}. On the left, we show the change point statistic for the Sinkhorn divergence with (SinkDivLM) and without (SinkDiv) a learned metric. 
    %The learned metric causes change statistic to be higher between samples across the change point, leading to improved performance. 
  To the right, we show Type 1 vs Type 2 errors for both approaches as we vary the (B) window sizes and (C) amount of added noise.}
    \label{Fig: GMM subplots}
 \end{figure*} 

\begin{table}[t]
  \caption{{\em AUC on high-dimensional datasets}. SinkDivLM is used with $\ell_1$-regularization to learn a sparse metric.}
  \centering
  \begin{tabular}{lcccc}
    \toprule
    Dataset  & HSIC & sHSIC & SinkDiv  & \textbf{SinkDivLM} \\
    \midrule
    Sleep Stage & 0.668 & 0.941 & 0.925 & 0.946  \\
    Swch Var & 0.868 & 0.934 &  0.567 & 0.931\\
    Swch Frq/Slp & 0.592 & 0.521 & 0.331  & 0.672\\
    \bottomrule
  \end{tabular}
  \label{Table: Sparse}
\end{table}
\subsection{Learning a sparse metric for feature selection}

When $\ell_1$ regularization is used, the learned ground metric can be used for selecting features that correspond to changes of interest. For the switching variance dataset, the learned metric $\mL$ is large in magnitude for indices corresponding to the dimension in which the variance is changing.  The true changes between REM and non-REM sleep  in the mouse sleep stage dataset also correspond to changes in only some of the 48 dimensions (different neurons), which the $\ell_1$ regularized ground metric is able to select.
Table \ref{Table: Sparse} shows results on switching variance and sleep stage datasets. As these datasets are high-dimensional, $\ell_1$ regularization is used to learn a sparse ground metric.

Though HSIC is presented in \cite{yamada2013change} as an unsupervised method that finds features that maximize separation by using pseudolabels at every time instance, we can also use HSIC for feature selection in a supervised manner by focusing on true change points. Both HSIC and our method aim to focus on finding a small number of features that can predict changes of interest. However, our method can also identify multivariate patterns (or correlations in different variables) that must be present for a CP to be detected. For this reason, when we tested both methods on the switching frequency with slopes dataset. HSIC fails to identify the correct feature in switching frequency dataset and mistakenly identifies other dimensions with constant slope as features causing the change. 
%HSIC looks for feature having the largest dependence to a change. The non-frequency changing features though have the same slope before and after the change, the signal, in terms of its values, is different before and after the change, making HSIC mistakenly identify this as a feature of interest.
Moreover, through using a triplet loss in our approach, we can identify features whose Sinkhorn divergence is smaller for sub-sequences before the change than sub-sequences across a change. This allows it to correctly identify the feature of interest causing the change. 

%\begin{figure}[h]%
%    \centering
%    \begin{subfigure}[Beedance sequence ]{
%        \centering
%        \label{fig: MG_ae_30_semi}%
%    \includegraphics[width=0.22\textwidth]{figures/bee_dance_example.pdf}
%    }\end{subfigure}
%    \begin{subfigure}[Learned metric]
%     {   \centering
%         \label{fig: Bee Dance metric}%
%         \includegraphics[width=0.22\textwidth]{figures/Beedance_metric.pdf}
%    }\end{subfigure}
%    \caption{Learned metric ($\mM = \mL^T\mL$) for Bee Dance dataset. The metric learned by our approach weighs  the third dimension heavily and uses the first two dimensions to a lesser extent.}
%    \label{}
%\end{figure} 
\label{sec: Supp_beedance}
\begin{figure}[t!]%
    \centering
    \begin{subfigure}[Learned metric for Bee Dance]{
        \centering
        \label{Fig: Beedance metric}%
        \includegraphics[width=0.30\textwidth]{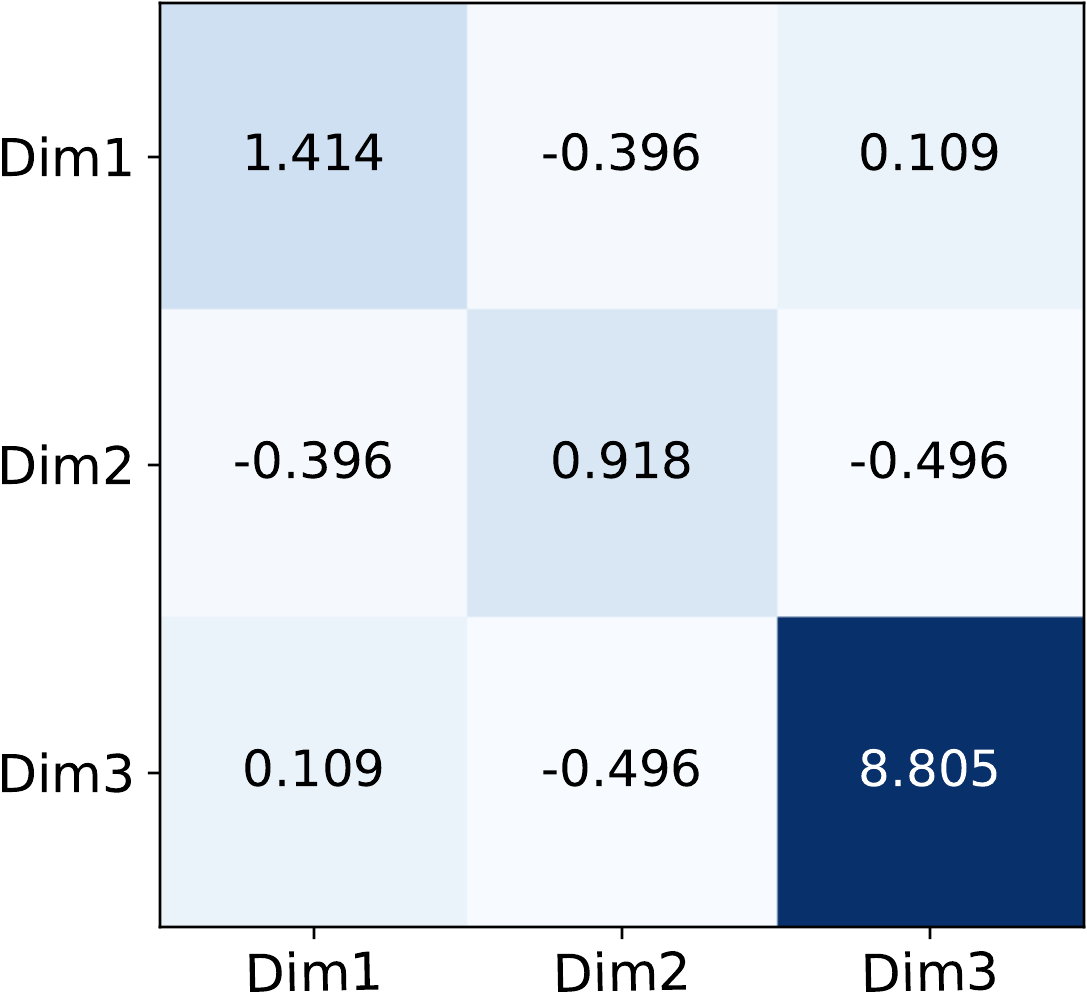}
    }\end{subfigure}
    \begin{subfigure}[Sample sequence from the Bee Dance dataset]{
        \centering
        \label{Fig: Beedance sequence}%
        \includegraphics[width=0.37\textwidth]{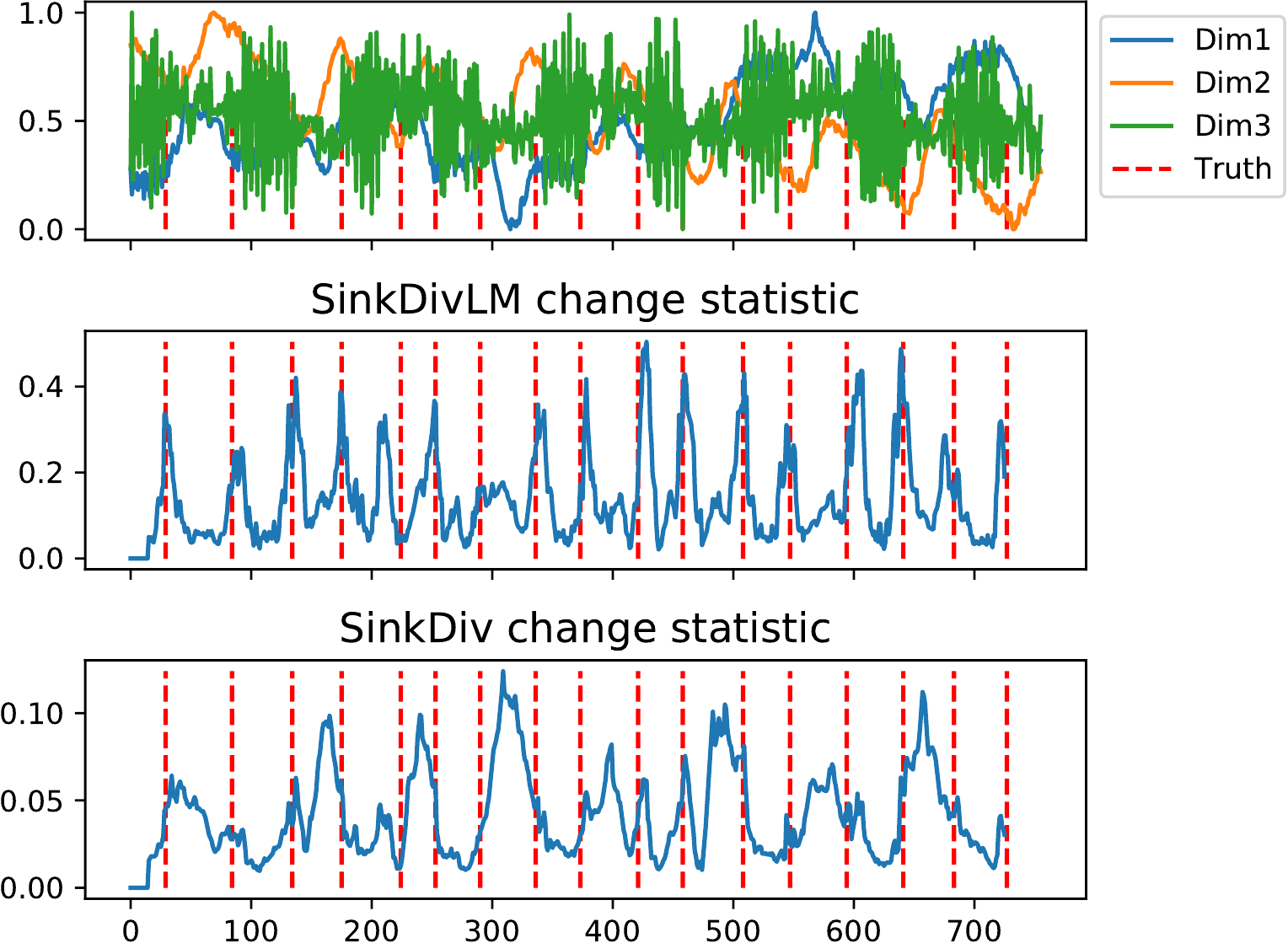}
    }\end{subfigure}
    \caption{ The first subplot in \ref{Fig: Beedance sequence}  shows a sample sequence from the Bee dance dataset. It can be seen that true changes, shown by red vertical lines, are often associated by changes in the variance of the third (green) dimension of this sequence. The learned metric in \ref{Fig: Beedance metric} captures this information as the 3 dimension is associated with a much larger value as compared to other dimensions. The 2nd subplot in \ref{Fig: Beedance sequence} shows that our method allows Sinkhorn divergences to use this learned metric to do a much better job at identifying change points than Sinkhorn divergences without a learned metric.}  
    \label{Fig: beedance interpet}
 \end{figure} 
\begin{figure}[t!]%
    \centering
    \begin{subfigure}[Learned metric for HASC]{
        \centering
        \label{Fig: HASC metric}%
        \includegraphics[width=0.30\textwidth]{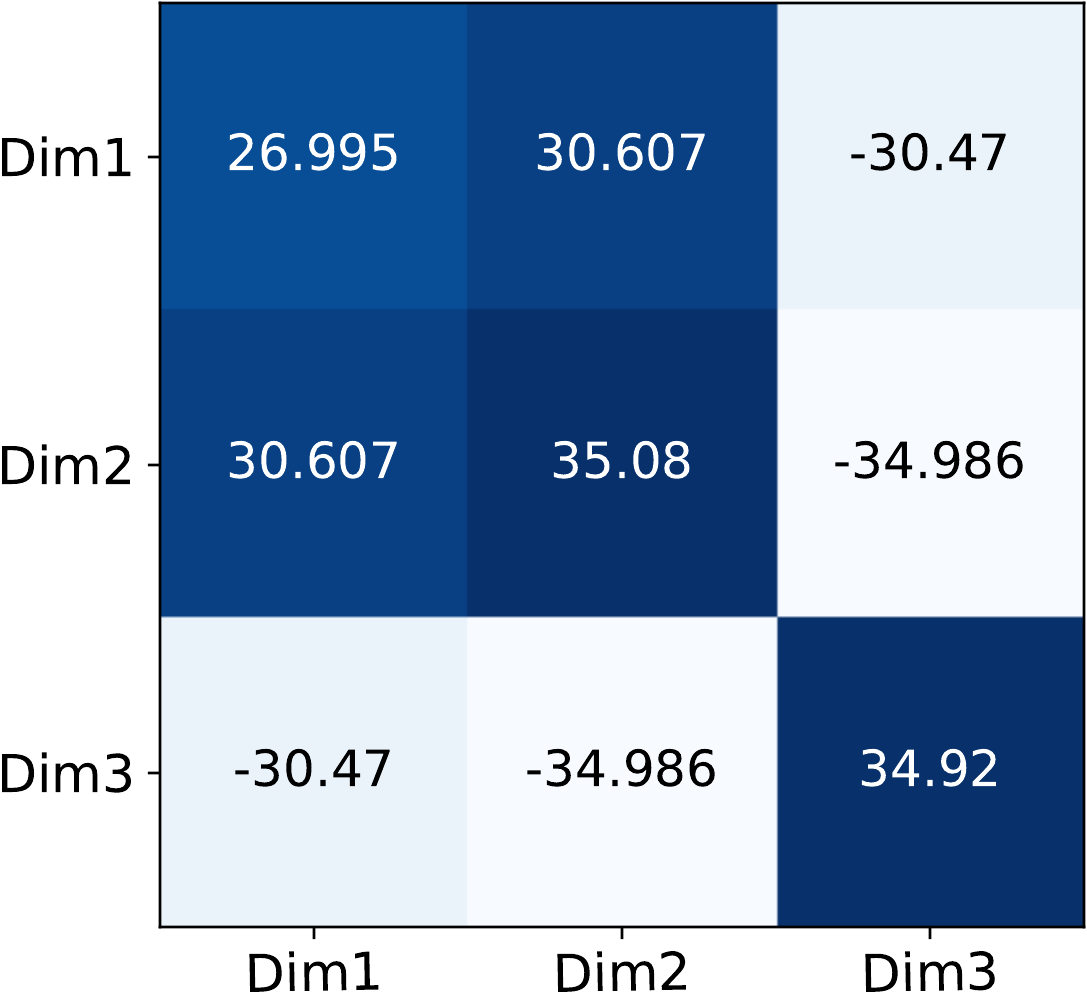}
    }\end{subfigure}
    \begin{subfigure}[Sample sequence from the HASC dataset ]{
        \centering
        \label{Fig: HASC sequence}%
        \includegraphics[width=0.37\textwidth]{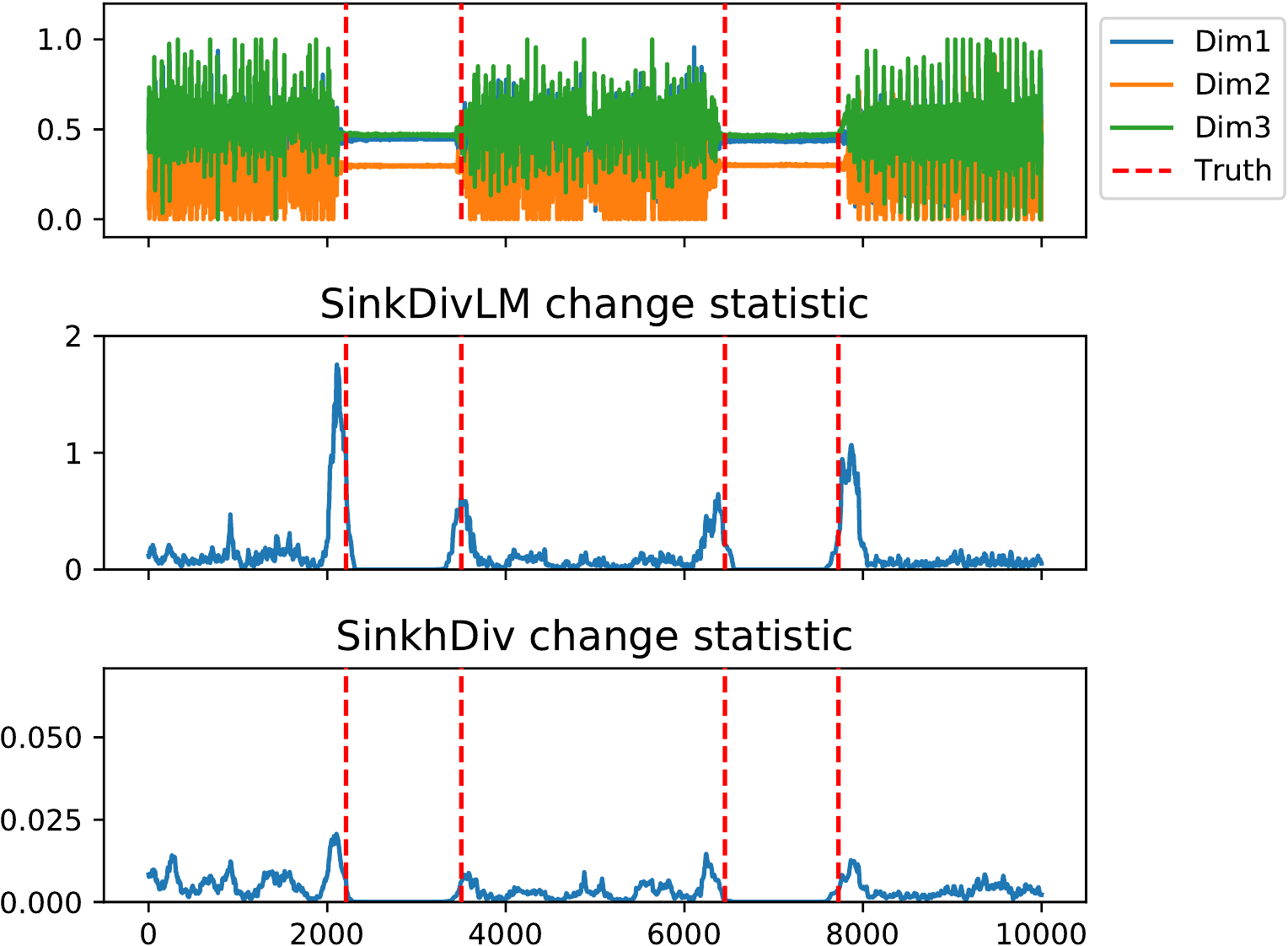}
    }\end{subfigure}
    \caption{\ref{Fig: HASC metric} shows the learned metric for the HASC dataset, while \ref{Fig: HASC sequence} shows an example sequence with true change points shown by vertical lines. The learned metric. As compared to Sinkhorn divergence without a learned metric (SinkDiv),  learned metric results (SinkDivLM) in higher change statistic between sub-sequences across true changes points (higher testing power), and lower test statistic in regions with no changes that leading to a lesser number of false change points.}
    \label{Fig: HASC}
 \end{figure} 
 \subsection{Interpretability}
A learned linear metric also allows us to interpret how different input features contribute to  detecting changes. For example,  Figure \ref{Fig: beedance interpet} shows that third dimension of the input sequence is mostly responsible for detecting changes. This helps us get a better understanding of what \emph{kinds} of changes in sequences are of interest.
For the Human activity dataset, the learned metric in Figure \ref{Fig: HASC} shows what input features are positively correlated and what features are negatively correlated. This learned metric increases the change statistic at true change points, leading to better change detection performance.
Figure \ref{Fig: Sleep stage plot} shows the sparse learned metric for the Sleep Stage dataset. This helps us identify what features, or neurons in the hippocampus, are responsible for causing changes between REM and non-REM sleep stages. Tables \ref{Table: Feat Neuro SinkDiv} and \ref{Table: Feat Neuro HSIC} show the top 5 neurons identified by SinkDivLM and HSIC respectively for causing changes between REM and Non-REM sleep stages. These neurons are visualized in Figures \ref{Fig: Slee Stage Identified SinkDiv} and \ref{Fig: Sleep stage Identified features HSIC}.
\label{sec: sleep stage}
\begin{figure}[t!]%
    \centering
    \begin{subfigure}[Sparse metric learned by SinkDivLM ofor Sleep Stage dataset]{
        \centering
        \label{Fig: Sparse HAS metric}%
        \includegraphics[width=0.60\textwidth]{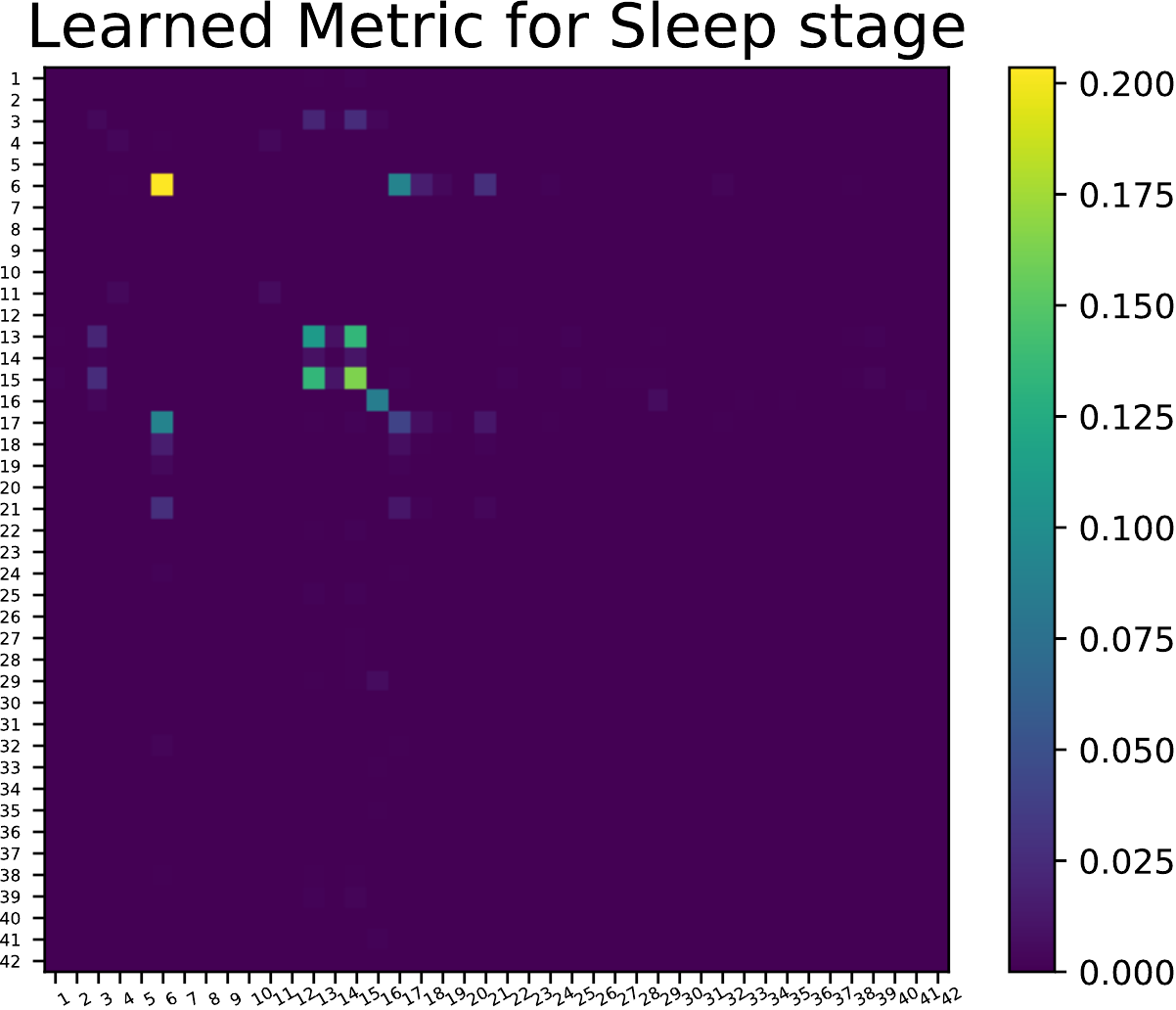}
    }\end{subfigure}
    \begin{subfigure}[Top 5 Neurons identified by SinkDivLM]{
        \centering
        \label{Fig: Slee Stage Identified SinkDiv}%
        \includegraphics[width=0.45\textwidth]{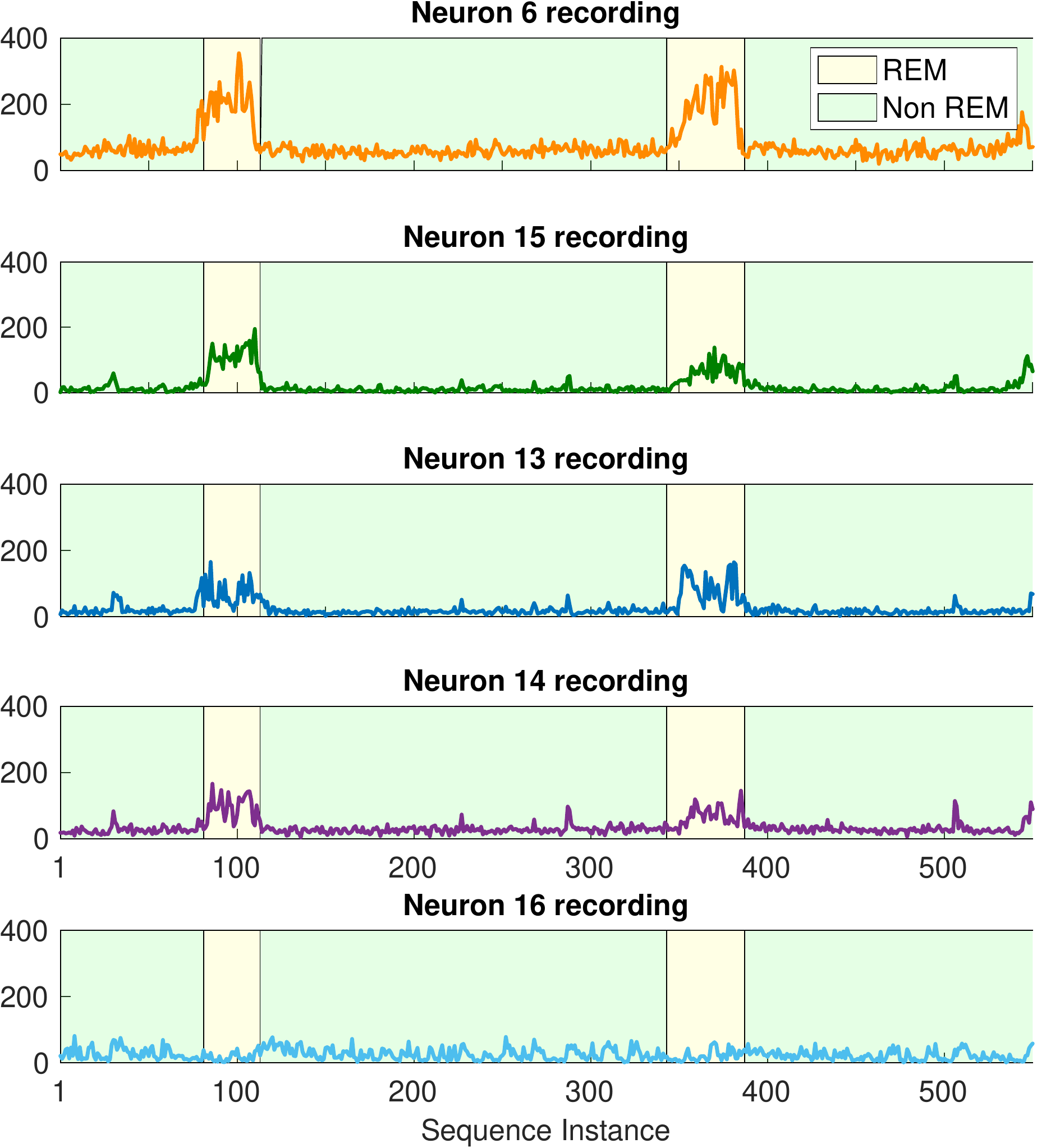}
    }\end{subfigure}
    \begin{subfigure}[Top 5 Neurons identified by sHSIC]{
        \centering
        \label{Fig: Sleep stage Identified features HSIC}%
        \includegraphics[width=0.45\textwidth]{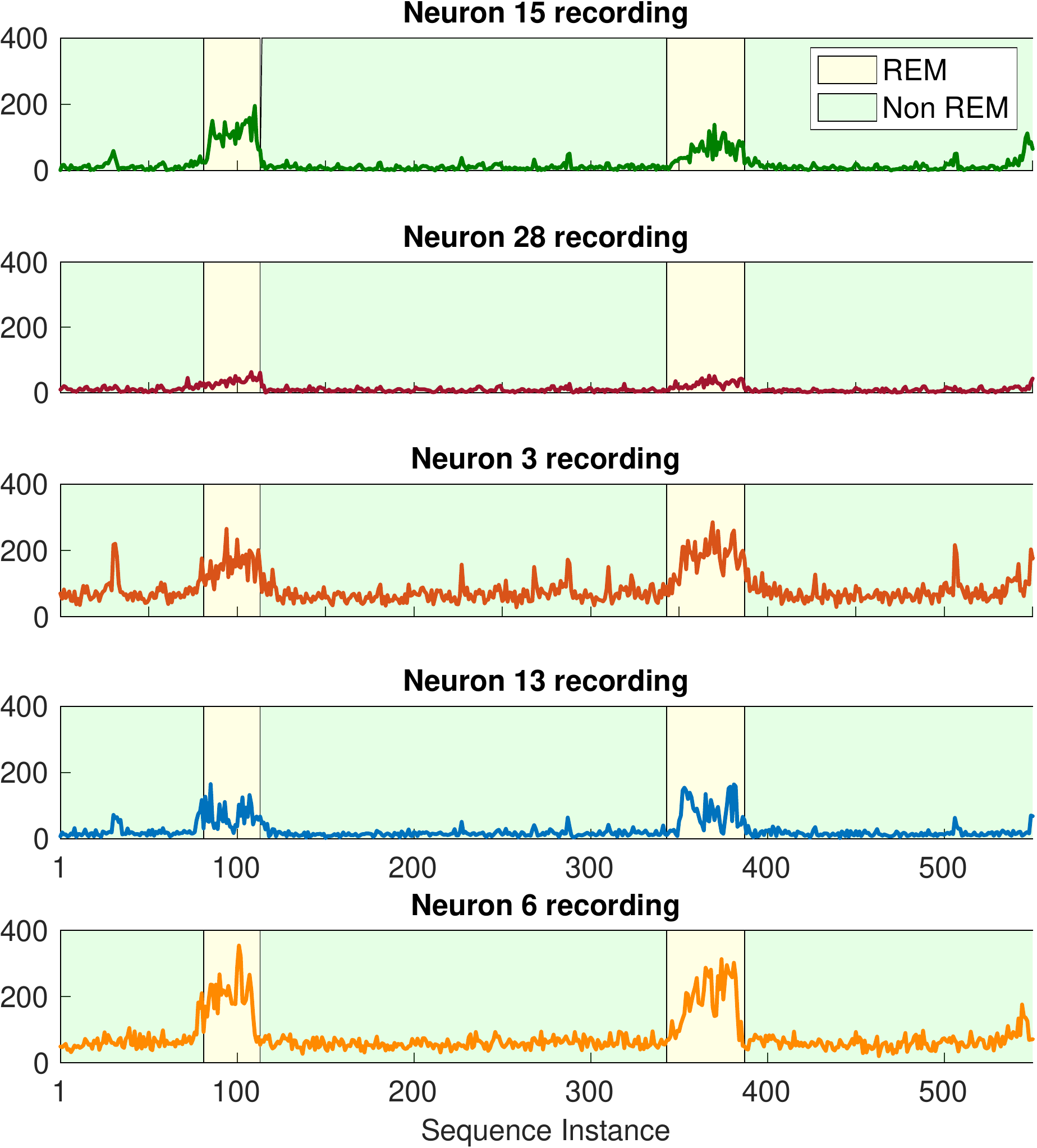}
    }\end{subfigure}
    \caption{\ref{Fig: Sparse HAS metric} shows the learned sparse metric for the Sleep Stage dataset. The top 5 features, or neurons, from this metric are  are visualized  in \ref{Fig: Slee Stage Identified SinkDiv}, while \ref{Fig: Sleep stage Identified features HSIC} visualizes top 5 features identified by sHSIC.}
    \label{Fig: Sleep stage plot}
 \end{figure}

\label{sec: Sup Error experiments}
\begin{figure}[t]%
    \centering
    \begin{subfigure}[Errors across different projection dimensions of $\mL$]{
        \centering
        \label{Fig: GMM dim}%
        \includegraphics[width=0.4\textwidth]{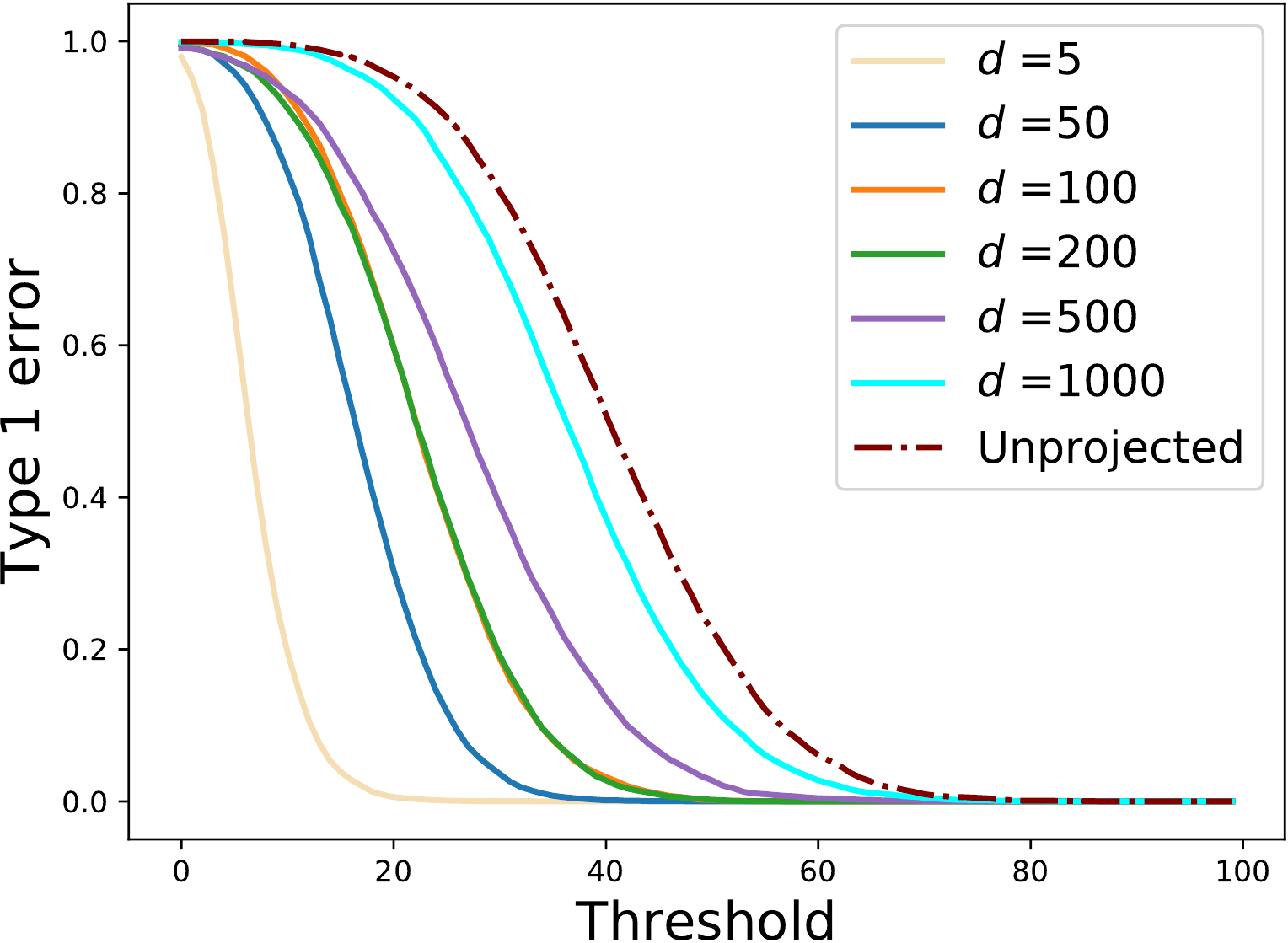}
    }\end{subfigure}
    \begin{subfigure}[Projection dimension vs rank of $\mL^T\mL$]{
        \centering
        \label{Fig: GMM effective rank}%
        \includegraphics[width=0.4\textwidth]{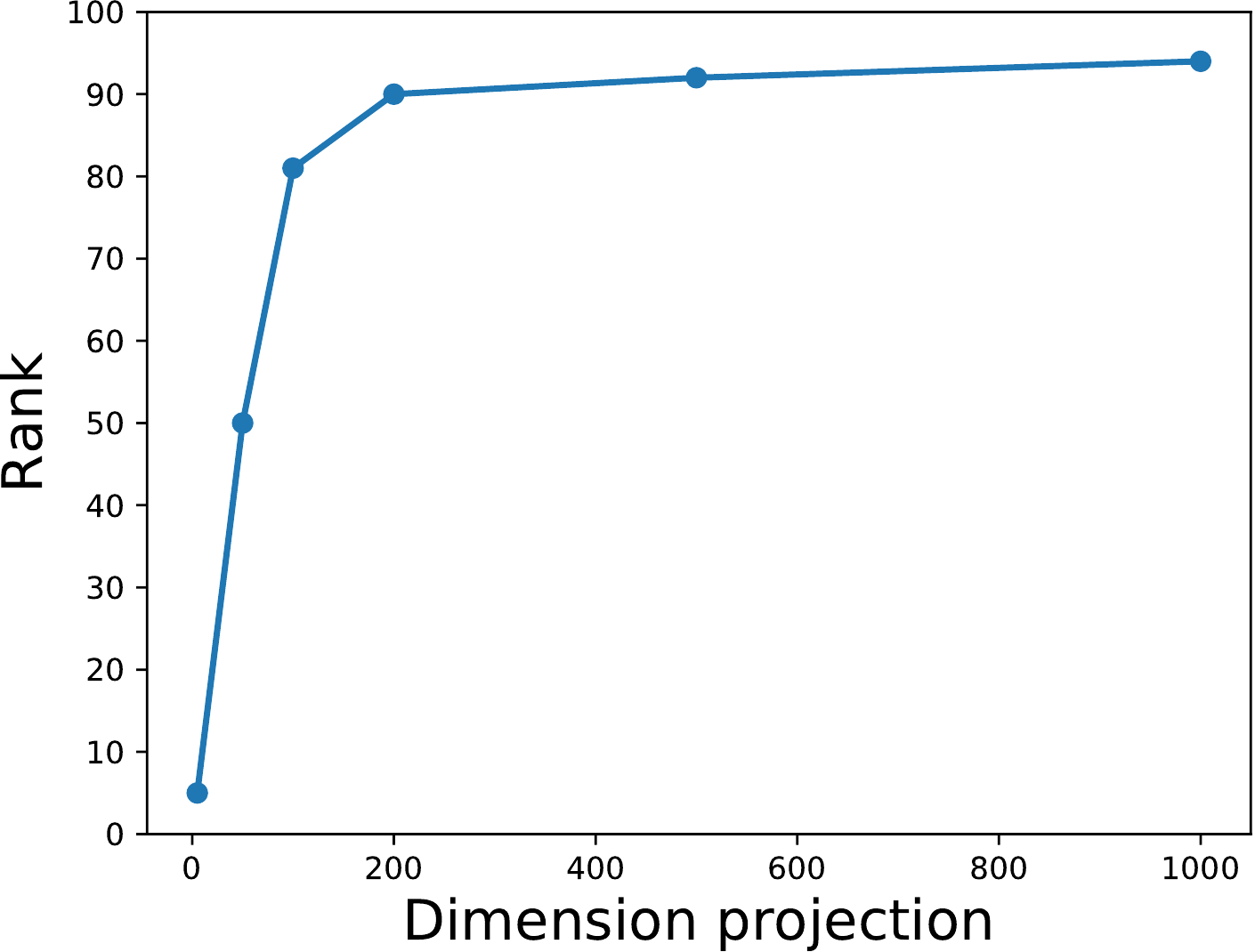}
    }\end{subfigure}
    \caption{\ref{Fig: GMM dim} shows the Type 1 errors for different dimension projections for the transformation $\mL$. \ref{Fig: GMM effective rank} shows the rank for the Mahalanobis metric ($\mL^T\mL$) that is induced by $\mL$. For dimension projections  greater than 50, the effective rank was lower than the ambient dimension, leading to better Type1 error rates than the unprotected case. This held true even when the projection dimension was 1000.}
    \label{Fig: GMM }
 \end{figure}

  \begin{table}[h!]
  \caption{Top 5 Neurons identified by SinkDivLM on Sleep Stage dataset}
  \centering
  \begin{tabular}{lccccc}
    \toprule
     & Neuron 6 & Neuron 15 & Neuron 13 & Neuron 14 & Neuron 16  \\
    \midrule
    Normalized Feature value & 1.00 & 0.8058 & 0.5405 & 0.4225 & 0.226  \\
    \bottomrule
  \end{tabular}
  \label{Table: Feat Neuro SinkDiv}
\end{table}
 \begin{table}[h!]
 \centering
  \caption{Top 5 Neurons identified by sHSIC on Sleep Stage dataset }
  \begin{tabular}{lccccc}
    \toprule
    & Neuron 15 & Neuron 28 & Neuron 3 & Neuron 13 & Neuron 6  \\
    \midrule
    Normalized Feature value & 1.00 & 0.464 & 0.452 & 0.436 & 0.384    \\
    \bottomrule
  \end{tabular}
  \label{Table: Feat Neuro HSIC}
\end{table}

\subsection{Type1 error versus projection dimension}
We further study the results in \ref{eq: Null} that relate how projection dimension of $\mL$ affects type 1 error. For this, we conducted two samples tests between samples from two 100 dimensional Gaussian mixture models; $\valpha = \mathcal{N}(\mathbf{0},\mI) + \mathcal{N}(\mathbf{1},\mathbf{\Sigma}_0) $ and samples from $\vbeta = \mathcal{N}(\mathbf{0},\mI) + \mathcal{N}(\mathbf{1.5},\mathbf{\Sigma}_1)$. $\mathbf{\Sigma}_0$ and $\mathbf{\Sigma}_1$ are diagonal covariance matrices where the first 3 entries on the diagonal are 3 and 5 respectively, while the rest of the diagonal entries are 1. 
$\valpha$ is chosen to be the null distribution and $\vbeta$ is used as the alternate distribution for these experiments. An additional noise of $\mathcal{N}(0,2)$ was added to these samples. Samples from these two distributions were used to minimize \eqref{eq: loss func}  with different choices of projection dimension $\mL$.
Figure \ref{Fig: GMM } shows the results for type 1 error across $\mL$ learned with different projection dimensions.

For dimension projections  greater than 50, the effective rank was lower than the  ambient dimension of the input data, leading to better Type1 error rates than the unprotected case. This held true even when the projection dimension was 1000. This is an interesting observation we would like to further analyze in the future. Though our loss function doesn't explicitly require the learned metric to be low rank, the learned metric through the triplet loss resulted in a metric that was lower rank (than the ambient dimension), leading to smaller Type 1 errors.
\pagebreak
\section{Conclusion and future directions}

There are numerous ways in which we can further improve our method. For training sequences, we only obtain similar dissimilar pairs from true change point labels. We can increase the number of training pairs by using the learned metric to detect change points on the training sequences. Falsely detected change points can be used to obtain similar pairs which could in turn be used to increase the number of similarity triplets for retraining the metric.

Sinkhorn divergences provide a powerful tool for distinguishing between samples. However, they do not naturally incorporate temporal structure into the cost. This means that Sinkhorn divergences can not, for example, distinguish a sequence from its temporally inverted counterpart. One potential way to address this weakness is to leverage tools from  order preserving Wasserstein distances in 
\cite{su2017order,su2019learning} that regularize the transport plan so that the temporal nature of the data is respected. This makes sense for comparing sequences which have  similar starting time, an assumption that is violated by consecutive sequences obtained through sliding windows. This causes many false change points at the boundary of these sliding windows.  A possible future step would be to look for Sinkhorn divergences that incorporate temporal nature of the data but can be used on consecutive sliding windows.

It would also be interesting to see how supervised change point detection could be used in conjunction with unsupervised change point detection to further improve performance. One straight forward  method is to use our method in a transductive manner by first using  available labels to  learn a supervised ground metric for detecting change points.  This metric can be used to be detect unlabeled change points for retraining the ground metric.  
%Other promising directions could be to reduce the number of true change points needed by using active learning based methods and learn a non-linear ground metric using neural networks. %However, such an approach might require a much larger number of change points which are often scarce in real-world settings.

%We have proposed a method that uses true change information to learn a ground metric for improved change point detection. While requiring only relatively few true change points for training, we have shown through examples on simulated as well as real-world datasets that our method can improve change point detection performance.  As some examples of change points are often available in many settings, we believe our proposed method has the potential to be widely applicable in the real-world.


\bibliography{main}
\bibliographystyle{plain}

%%%%%%%%%%%%%%%%%%%%%%%%%%%%%%%%%%%%%%%%%%%%%%%%%%%%%%%%%%%%%%%%%%%%%%%%%%%%%%%
%%%%%%%%%%%%%%%%%%%%%%%%%%%%%%%%%%%%%%%%%%%%%%%%%%%%%%%%%%%%%%%%%%%%%%%%%%%%%%%
% APPENDIX
%%%%%%%%%%%%%%%%%%%%%%%%%%%%%%%%%%%%%%%%%%%%%%%%%%%%%%%%%%%%%%%%%%%%%%%%%%%%%%%
%%%%%%%%%%%%%%%%%%%%%%%%%%%%%%%%%%%%%%%%%%%%%%%%%%%%%%%%%%%%%%%%%%%%%%%%%%%%%%%
\newpage
\appendix
\onecolumn

\section{Data sources}

\textbf{Beedance:} \url{https://sites.cc.gatech.edu/~borg/ijcv_psslds/}\\
More details on the bee waggle dance can be seen at 
 \url{https://www.youtube.com/watch?v=1MX2WN-7Xzc}\\
\textbf{HASC}: \url{http://hub.hasc.jp}\\
\textbf{ECG}: \url{https://timeseriesclassification.com/description.php?Dataset=ECG200}\\
These sources can also be downloaded from datapages in:
\url{https://github.com/OctoberChang/klcpd_code}\\
\url{https://github.com/kevin-c-cheng/OtChangePointDetection}\\
\textbf{Sleep Stage}:   Extracellular single unit spiking was collected from chronically implanted, freely behaving animals \cite{su2017order,hengen2016neuronal}. Tetrode arrays were implanted without drives into mouse CA1 (C57BL/6) and rat V1 (Long Evans). Following recovery, neural data were recorded at 25 kHz continuously during free behavior. Raw data were processed and clustered using standard pipelines. Data was bandpassed (500-10,000 Hz) and clustered using MountainSort. Single units were identified in the clustering output via XGBoost.
Trained human scorers evaluated the LFP power spectral density and integral of animal movement to evaluate waking, NREM and REM sleep.
\section{Baseline sources}

\textbf{HSIC}: \url{https://github.com/riken-aip/pyHSICLasso}\\
\textbf{KLCPD}: \url{https://github.com/OctoberChang/klcpd_code}\\
\textbf{TIRE}: \url{https://github.com/deryckt/TIRE}

\section{Experiment details}
\begin{table}[H]
  \caption{Parameter settings for experiments}
  \centering
  \begin{tabular}{lccccc}
    \toprule
    Dataset &  Project dim ($d$) & Win size $w$ &  Entrp Reg ($\gamma$) & Learn rate ($\mu$) & $\ell_1$ Regul ($\lambda$) \\
    \midrule
    GMM switch & 5 & 10 & 0.1 & 0.01 \\
    Freq switch & 50 & 100 & 1 & 0.01 & \\
    Freq switch w slope & 50 & 100 & 1 & 0.01& 5e-5 \\
    Beedance & 3 & 15 & 0.1 & 0.01 \\
    HASC & 3 & 200 & 0.1 & 0.01 \\
    Yahoo & 5 & 2 & 0.1 & 0.001\\
    ECG  & 2 & 3 & 0.001 & 0.001\\
    Sleep stage & 42 & 15 & 1 & 0.01 & 0.01\\
    \bottomrule
  \end{tabular}
  \label{Table: Experiment parameters}
\end{table}

For all experiments, a total of 2000 iterations were used and the model with best validation loss was saved. 
For baselines that involved two-sample tests (such as M-Stats, KLCPD, SinkDiv), the same window sizes were used. For TIRE different window sizes were used till the best performance was attained. 
We do not use a window margin (where a change is correctly detected if it is within a certain margin of the true change point)

\vspace{4mm}

For time efficiency, sliding windows were used to obtain batched batched two-sample tests. two-sample tests using Sinkhorn divergence libraries for Pytorch were conducted on these batches. 

\textbf{Sinkhorn library}:\url{https://github.com/gpeyre/SinkhornAutoDiff}

\vspace{4mm}

For the Sleep stage dataset, the true change points between REM and non-REM sleep stages are often not labelled perfectly (There might not be any  prominent change at a true labelled change point for very short window sizes). For these reasons, when learning features through both sHSIC and SinkDivLM,  we select windows on the opposite side of change points with a buffer of size 10. This buffer is not needed when detecting change points over sliding windows.

\section{Alternate formulation}

Unfortunately, the loss function  in \eqref{eq: loss func} is not convex in $\mL$ as its Hessian, with respect to $\mL$, is not guaranteed to be positive semi-definite. Sinkhorn divergence with parameterized ground metric in \eqref{eq: OT plan parameterized} can be equivalently expressed as
\begin{align*}
    \mathcal{W}_{\mM,\gamma} (\mX, \mY ) 
=  \min_{\mP } &  \sum_{i=1}^n \sum_{j=1}^m \mP_{i,j}(\vx_i - \vy_j)^T\mM(\vx_i - \vy_j) - \gamma \mH(\mP) \\ \text{ subject to }  
    &\mP  \in \R_{+}^{n \times m} \nonumber \\ 
    &\mP^T\mathbbm{1}_n = 1 , \mP\mathbbm{1}_m = 1 \nonumber,
\end{align*} 

where $\mM = \mL^T\mL$. Consequently, $\mM$ can be learned by
%\[
%    l(\mM) = \sum_{i \in \text{Trip pairs}}  %\mathcal{S}_{\mM,\gamma} (\mX_i, \mX_{i}^s ) +%
%    [c - \left(\mathcal{S}_{\mM,\gamma} (\vx_i, \vx_{i_d}) - \mathcal{S}_{\mM,\gamma} \left(\vx_i, \vx_{i_s}\right) )\right  . ]^+ 
%\]
\begin{align*}
    &\min_{\mM}  \sum_{i \in \text{Triplets}}  %\mathcal{S}_{\mL,\gamma} (\mX_i, \mX_{i}^s ) + %
    \left[c - (\mathcal{S}_{\mM,\gamma} (\mX_i, \mX_{i}^d) - \mathcal{S}_{\mM,\gamma} \left(\mX_i, \mX_{i}^s\right) )\right   ]^+  \\ 
    &\text{ subject to }  
    \mM \succcurlyeq 0 \nonumber
\end{align*} 
The positive semi-definite condition on $\mM$ arises from the requirement on Wasserstein distance $\mathcal{W}_{\mM,\gamma}$, and subsequently $\mathcal{S}_{\mM,\gamma}$, to be positive. Additionally, $\mathcal{S}_{\mM,\gamma}$ is linear in $\mM$.

\section{Obtaining the transport plan for Sinkhorn distances}

The transport plan $\mP$ for the regularized Wasserstein distances (or Sinkhorn distances), can be obtained using the Sinkhorn algorithm. We first set up the dual formulation of \eqref{eq: reg OT}
\subsection{Dual formulation}
We can incorporate the constraints into a Lagrangian dual function
\begin{align}
     \max_{\mathbf{f}, \mathbf{g}} \min_{\mathbf{P}} L(\mathbf{P}, \mathbf{f}, \mathbf{g}) &= \max_{\mathbf{f}, \mathbf{g}} \min_{\mathbf{P}} \langle \mathbf{C} , \mathbf{P} \rangle  - \gamma E(\mathbf{P} ) +  \langle \mathbf{f} , 
     \mathbf{a} - \mathbf{P} \mathbbm{1}_m  \rangle + \langle \mathbf{g} , 
      \mathbf{b} - \mathbf{P}^T \mathbbm{1}_n  \rangle \nonumber \\
      &= \max_{\mathbf{f}, \mathbf{g}} \langle \mathbf{f}, \mathbf{a} \rangle + \langle \mathbf{g}, \mathbf{b} \rangle +  \min_{\mathbf{P}} \langle \mathbf{C} - \mathbf{f}\mathbbm{1}_m  - \mathbf{g}\mathbbm{1}_m, \mathbf{P} \rangle -\gamma E(\mathbf{P}) \nonumber \\
      &= \max_{\mathbf{f}, \mathbf{g}} \langle \mathbf{f}, \mathbf{a} \rangle + \langle \mathbf{g}, \mathbf{b} \rangle +  \min_{\mathbf{P}} \langle \mathbf{C} - \mathbf{f}\mathbbm{1}_m  - \mathbf{g}\mathbbm{1}_m, \mathbf{P} \rangle -\gamma \langle \mathbf{P} , \log \mathbf{P} - \mathbbm{1}_{n \times m} \rangle
      \label{eq: Sol Lang}
\end{align}

By solving $\frac{\partial  L(\mathbf{P}, \mathbf{f}, \mathbf{g}) }{\partial \mathbf{P}} = 0$, we can obtain $\mathbf{P}$ such that:
\begin{equation*}
    \mathbf{P}_{i,j} = e^{{\mathbf{f}_i}/\gamma }\underbrace{e^{{\mathbf{-C}_{i,j}/\gamma}}}_{\text{Kernel }} e^{{\mathbf{g}_j}/\gamma}
\end{equation*}

Substituting $\mathbf{P}$ in \eqref{eq: Sol Lang}, we can obtain after simplification the equivalent dual problem

\begin{equation}
\max_{\mathbf{f}, \mathbf{g}} \langle \mathbf{f}, \mathbf{a} \rangle +  \langle \mathbf{g}, \mathbf{n} \rangle -\gamma \langle e^{{\mathbf{f}}/\gamma}, e^{{\mathbf{-C}/\gamma}} e^{{\mathbf{g}}/\gamma}. \rangle 
\end{equation}   

For non-discrete distributions, a more generalized dual formulation can be seen below
\begin{align}
    \mathcal{W}_{\gamma}^p (\mathbf{a}, \mathbf{b} ) &=  \sup_{( f(x) \in \mathcal{C(X)} , g(y) \in \mathcal{C(Y)} )  } \int_\mathcal{X} f(x) d\alpha(x) + \int_\mathcal{Y} g(y)d\beta(y) - \gamma\int_{\mathcal{X,Y}} e^{\frac{f(x) + g(y) - c(x,y)^p}{\gamma}}d\alpha(x)d\beta(y) \\
    &= \sup_{( f(x) \in \mathcal{C(X)} , g(y) \in \mathcal{C(Y)} )} \mathbb{E}_{\alpha \beta}[z_{\epsilon}^{X,Y} (f,g)]  
\end{align}

where $z_{\epsilon}^{x,y} (f,g) = f(x) + g(y) - \gamma e^{\frac{f(x) + g(y) - c(x,y)^p}{\gamma}}$.

\subsection{Sinkhorn Algorithm}

We can rewrite \eqref{eq: Sol Lang} in terms of vectors $\mathbf{u}, \mathbf{u}$ and $\mathbf{K} = e^{{\mathbf{-C}_{i,j}/\gamma}} $ as:

\begin{equation}
    \mathbf{P} = \text{diag}(\mathbf{u} ) \mathbf{K} \text{diag}(\mathbf{v} )
\end{equation}

Also from constraints: \[ \text{diag}(\mathbf{u} ) \mathbf{K} \text{diag}(\mathbf{v} ) \mathbbm{1}_m = \mathbf{a}  \text{    and   } \left( \text{diag}\mathbf{u} ) \mathbf{K} \text{diag}(\mathbf{v} ) \right)^T \mathbbm{1}_n = \mathbf{b} , \]

\[ \mathbf{u} \odot \mathbf{K} \mathbf{v} = \mathbf{a } \text{    and   }    \mathbf{v} \odot \mathbf{K}^T \mathbf{u} = \mathbf{b } . \]

An alternating update  scheme can be used to update the dual potentials until convergence
\[ \mathbf{u} ^{l + 1} = \frac{\mathbf{a}}{\mathbf{K}\mathbf{v}^l} \text{     and   }   \mathbf{v} ^{l + 1} = \frac{\mathbf{b}}{\mathbf{K}^T\mathbf{u}^l}.   \]

\end{document}